%% file: tech_main.tex
\documentclass[oneside,a4paper,onecolumn,11pt]{article}

\usepackage{fullpage}
\usepackage[left=2cm,top=2cm,bottom=2cm,right=2cm,includehead,nomarginpar,headheight=16pt]{geometry}
\usepackage[ruled, linesnumbered, vlined, commentsnumbered]{algorithm2e}
\usepackage{graphicx,subfig} 
\usepackage{amsfonts,amssymb,amsmath,amsthm,amsopn,mathtools}	
\usepackage{booktabs,diagbox,colortbl,multirow,tabularx,threeparttable,hhline}
\usepackage[listings,skins,breakable]{tcolorbox}

\usepackage{fancyhdr,fancyheadings,nopageno,lastpage} 
\usepackage{url}
\usepackage{enumerate}
\usepackage[shortlabels]{enumitem}
\usepackage{csquotes}
\usepackage{authblk}
\usepackage{footnote}
\usepackage{hyperref}
\usepackage{prettyref}
\usepackage{cite}
\usepackage{setspace}
\usepackage{color}
\usepackage{xcolor}  
\usepackage{geometry}
\usepackage{academicons}
\usepackage{fontawesome5}
\usepackage{pifont,ifsym,marvosym,manfnt} 

\usepackage{tikz}
\usepackage{pgfplots}
\usetikzlibrary{positioning,shapes,shadows,arrows,calc}
\tikzstyle{component}=[rectangle, draw=black, rounded corners, fill=blue!40, drop shadow, text centered, anchor=north, text=white, minimum height=1cm]
\tikzstyle{arrow}=[->, thick]

\pgfplotsset{compat=1.12}
\usetikzlibrary{intersections}
\usetikzlibrary{pgfplots.statistics}
\usepgfplotslibrary{fillbetween}

\fancypagestyle{plain}{%
    \fancyfoot[C]{}
    \fancyfoot[R]{\faLeanpub\ \, \thepage\ / \pageref{LastPage}}
    \fancyfoot[L]{\faBraille\ \textsc{COLALab Report} \faSlackHash\ \footnotesize\textifsym{2024003}}
}

\hypersetup{
    colorlinks=true,
    linkcolor=ultramarine,
    filecolor=magenta,
    urlcolor=ultramarine,
    pdftitle={Overleaf Example},
    pdfpagemode=FullScreen,
}

\definecolor{red(munsell)}{rgb}{0.95, 0.0, 0.24}
\definecolor{navyblue}{RGB}{0, 0, 128}
\definecolor{myblue}{RGB}{34,31,217}
\definecolor{mycyan}{gray}{.7}
\definecolor{Gray}{gray}{0.9}
\definecolor{usccardinal}{rgb}{0.6, 0.0, 0.0}
\definecolor{ultramarine}{RGB}{0,32,96}
\definecolor{amber}{rgb}{1.0, 0.49, 0.0}

\newtheorem{remark}{Remark}

\newtheorem{definition}{Definition}

\newtcolorbox{quotebox}{colback=gray!10,boxrule=0.4pt,colframe=black,fonttitle=\bfseries,top=1pt,bottom=1pt}

\usepackage[ruled, linesnumbered, vlined, commentsnumbered]{algorithm2e}
\usepackage{prettyref}
\usepackage{amsfonts,amssymb,amsmath,amsthm,amsopn,mathrsfs,mathtools,wasysym}	
\usepackage{bm}
\usepackage{multirow}

\def\our{\texttt{CBOB}}

\SetCommentSty{mycommfont}

\DeclareMathOperator*{\argmax}{argmax}
\DeclareMathOperator*{\argmin}{argmin}

\newcommand{\pref}{\prettyref}

\newrefformat{fig}{Fig.~\ref{#1}}
\newrefformat{tab}{Table~\ref{#1}}
\newrefformat{sec}{Section~\ref{#1}}
\newrefformat{alg}{Algorithm~\ref{#1}}
\newrefformat{property}{Property~\ref{#1}}
\newrefformat{theorem}{Theorem~\ref{#1}}
\newrefformat{definition}{Definition~\ref{#1}}
\newrefformat{corollary}{Corollary~\ref{#1}}
\newrefformat{lemma}{Lemma~\ref{#1}}
\newrefformat{conj}{Conjecture~\ref{#1}}
\newrefformat{def}{Definition~\ref{#1}}
\newrefformat{eq}{equation~(\ref{#1})}
\newrefformat{app}{Appendix~\ref{#1}}

\usepackage{lscape}

\newenvironment{code-example}
{
\vspace{0.15cm}
\noindent\begin{minipage}{\linewidth}
\begin{center}
\arrayrulecolor{black}
\color{black}
\begin{tabular}{|p{0.95\linewidth}|}
\hline%
\rowcolor{pink!20}%
}
{
\\\hline
\end{tabular}
\end{center}
\end{minipage}
\vspace{-0.2cm}
}

\begin{document}

\title{\vspace{-1ex}\LARGE\textbf{Human-in-the-Loop Policy Optimization for Preference-Based Multi-Objective Reinforcement Learning}}

\author[1]{\normalsize Ke Li}
\author[2]{\normalsize Han Guo}
\affil[1]{\normalsize Department of Computer Science, University of Exeter, EX4 4RN, Exeter, UK}
\affil[2]{\normalsize School of Computer Science and Engineering, University of Electronic Science and Technology of China, Chengdu 611731, China}
\affil[\Faxmachine\ ]{\normalsize \texttt{k.li@exeter.ac.uk}}

\date{}
\maketitle

\vspace{-3ex}
{\normalsize\textbf{Abstract: }}Multi-objective reinforcement learning (MORL) aims to find a set of high-performing and diverse policies that address trade-offs between multiple conflicting objectives. However, in practice, decision makers (DMs) often deploy only one or a limited number of trade-off policies. Providing too many diversified trade-off policies to the DM not only significantly increases their workload but also introduces noise in multi-criterion decision-making. With this in mind, we propose a human-in-the-loop policy optimization framework for preference-based MORL that interactively identifies policies of interest. Our method proactively learns the DM's implicit preference information without requiring any a priori knowledge, which is often unavailable in real-world black-box decision scenarios. The learned preference information is used to progressively guide policy optimization towards policies of interest. We evaluate our approach against three conventional MORL algorithms that do not consider preference information and four state-of-the-art preference-based MORL algorithms on two MORL environments for robot control and smart grid management. Experimental results fully demonstrate the effectiveness of our proposed method in comparison to the other peer algorithms.

{\normalsize\textbf{Keywords: }}Multi-objective reinforcement learning (MORL), human-in-the-loop, preference learning, decomposition multi-objective optimization.

\input{introduction}

\input{preliminary}

\input{method}

\input{settings}

\input{experiment}

\input{conclusion}

\section*{Acknowledgment}
This work was supported in part by the UKRI Future Leaders Fellowship under Grant MR/S017062/1 and MR/X011135/1; in part by NSFC under Grant 62376056 and 62076056; in part by the Royal Society under Grant IES/R2/212077; in part by the EPSRC under Grant 2404317; in part by the Kan Tong Po Fellowship (KTP\textbackslash R1\textbackslash 231017); and in part by the Amazon Research Award and Alan Turing Fellowship.

\bibliographystyle{IEEEtran}
\bibliography{IEEEabrv,pbmorl}

\newpage
\input{appendix}

\end{document}

%% file: introduction.tex

\section{Introduction}
\label{sec:introduction}

Many real-world decision-making tasks involve more than one potentially competing objectives. For example, the trade-offs between makespan and energy consumption in workflow scheduling~\cite{QinWYLZ20}, the balance between microgrid benefits and power grid requirements in microgrid system development~\cite{XuLA21}, and the competing considerations of fuel cost, efficiency, and safety in robot control~\cite{XuTMRSM20}. Multi-objective reinforcement learning (MORL) algorithms have gained increasing attention as a powerful paradigm for learning control policies that optimize over multiple conflicting criteria simultaneously~\cite{HayesRBKMRVZDHH22}. The existing MORL literature can be broadly classified into three main categories.

The first and most prevalent category of MORL approaches involves aggregating multiple objective functions into a single scalar reward using a convex combination, a technique commonly referred to as linear scalarization, as exemplified by~\cite{GaborKS98,MannorS01}. Specifically, the weight assigned to each objective represents the decision maker's relative preference for that particular objective. Subsequently, a standard reinforcement learning (RL) algorithm is employed to optimize this scalar reward. While this approach is intuitive, its broader applicability is constrained by the need for \textit{a priori} preferences regarding the different criteria in the underlying MORL task. Obtaining such preferences is often unrealistic in real-world black-box scenarios.

Another widely adopted strategy in MORL involves identifying a set of Pareto-optimal policies that span the entire spectrum of possible preferences in the domain, which are then presented to the decision maker (DM) to choose the policies that align with their interests, as done in~\cite{NatarajanT05,YangSN19,IkenagaA18}. This \textit{a posteriori} decision-making approach, however, not only increases the DM's workload but also introduces potentially irrelevant or even noisy information during the multi-criterion decision-making (MCDM) process. Moreover, it is not guaranteed to yield a diverse set of trade-off solutions that adequately cover the entire Pareto front within a limited budget, thereby potentially hindering the discovery of policies of interest in MCDM.

Distinct from the previous two categories of approaches, the last one is \textit{interactive} MORL paradigm, an emerging area first introduced in~\cite{VamplewDBID11}. It incorporates human-in-the-loop interaction in MORL, enabling DMs to progressively learn the characteristics of the underlying task. Consequently, the MORL process is \lq supervised\rq\ to identify policies of interest. As the DM has more frequent opportunities to provide new information, they may feel more in control and engaged in the overall \textit{optimization-cum-decision-making} process. Surprisingly, this line of research has received relatively limited attention in the current MORL community, as highlighted by~\cite{RoijersZN17,RoijersZLLPNA18,PeschlZOS22}.

In this paper, we present a general framework (dubbed \our) for designing interactive MORL algorithms that involve human-in-the-loop interaction, enabling the progressive learning of the DM's feedback and the adaptation of the search for policies of interest. This framework comprises three key modules: \texttt{seeding}, \texttt{preference elicitation}, and \texttt{optimization}.
\begin{itemize}
    \item The \texttt{seeding} module functions as an initial process to search for a set of trade-off policies that offer a reasonable approximation of the Pareto-optimal front.
    \item The \texttt{preference elicitation} module focuses on gathering the DM's preference information and converting it into a format that can be utilized in the subsequent \texttt{optimization} module.
    \item The \texttt{optimization} module encompasses any policy optimization algorithm used in deep RL. It employs the elicited preference information to guide the search for preferred policies in a parallel manner.
\end{itemize}
In practice, our proposed \our\ framework is highly versatile and readily applicable to complex environments featuring high-dimensional and continuous state and action spaces. Moreover, its modular design allows for flexibility, as each module can be adapted to other dedicated configurations. We conducted extensive experiments on two sets of MORL benchmark problems based on \textsc{MuJoCo}~\cite{TodorovET12} and a multi-microgrid system design problem~\cite{ChiuSP15}, which thoroughly demonstrate the effectiveness and competitiveness of our proposed \our\ algorithm instance in comparison to three conventional MORL and four state-of-the-art preference-based MORL peer algorithms.

In the rest of this paper, we will provide some preliminary knowledge pertinent to this paper along with a pragmatic review of existing works on constrained multi-objective optimization in~\pref{sec:preliminaries}. In~\pref{sec:method}, we will delineate the algorithmic implementations of our proposed \our\ framework. The experimental settings are given in~\pref{sec:settings}, and the empirical results are presented and discussed in~\pref{sec:results}. At the end, \pref{sec:conclusion} concludes this paper and sheds some lights on future directions.

%% file: preliminary.tex

\section{Preliminary}
\label{sec:preliminaries}

This section starts with some basic definitions pertinent to this paper, followed by a pragmatic overview of some selected up-to-date development of MORL and preference learning.

\subsection{Multi-Objective Markov Decision Process}
\label{sec:momdp}

In this paper, a MORL is formulated as a $5$-tuple $\langle \mathcal{S},\mathcal{A},T,\mathbf{r},\boldsymbol{\gamma}\rangle$ multi-objective Markov decision process (MOMDP):
\begin{itemize}
    \item $\mathcal{S}$ is the state space, i.e., the set of all available $n$-dimensional states $\mathbf{s}=(s_1,\ldots,s_n)^\top\in\mathbb{R}^n$.

    \item $\mathcal{A}$ is the action space, i.e., the set of all available $l$-dimensional actions $\mathbf{a}=(a_1,\ldots,a_l)^\top\in\mathbb{R}^l$.

    \item $T$ is the transition matrix where $T(s_{t+1}|s_t,a_t)$ represents the probability that the state $s_t$ transfers to the other $s_{t+1}$ after taking the action $a_t$.

    \item $\mathbf{r}=\left(r_1(s_t,a_t),\ldots,r_m(s_t,a_t)\right)^\top$ is the reward vector after taking the action $a_t$.

    \item $\boldsymbol{\gamma}=(\gamma_1,\ldots,\gamma_m)^\top\in(0,1]^m$ is a vector of discount factors.
\end{itemize}
In a MOMDP, a policy $\pi:\mathcal{S}\rightarrow\mathcal{A}$ determines how the current state $s_t$ move to the next one $s_{t+1}$ by taking the action $a_t\sim\pi(s_t)$. In particular, $\pi$ is associated with a vector of expected returns $\mathbf{J}(\pi)=\left(J_1(\pi),\cdots,J_m(\pi)\right)^\top$ where $J_i(\pi)=\mathbb{E}\left(\sum_{t=0}^H\boldsymbol{\gamma}_i^tr_i(s_t,a_t)\right)$, and $H>0$ is the horizon of the MOMDP.

\subsection{Multi-Objective Optimization}
\label{sec:mo}

The multi-objective optimization (MOP) problem considered in this paper is defined as:
\begin{equation}
    \begin{array}{l l}
        \mathrm{maximize}\quad\mathbf{F}(\pi)=\left(f_{1}(\pi),\cdots,f_{m}(\pi)\right)^\top\\
        \mathrm{subject\ to} \;\;\; \pi:\mathcal{S}\rightarrow\mathcal{A}
    \end{array},
    \label{eq:MOP}
\end{equation}
where $\pi$ is a policy and $\mathbf{F}(\pi)=\mathbf{J}(\pi)$ is an objective vector. Note that all objective functions are supposed to be conflicting with each other. That is to say the improvement of one objective can lead to the detriment of the others.

\begin{definition}
    Given two policies $\pi^1$ and $\pi^2$, $\pi^1$ is said to \underline{dominate} $\pi^2$ (denoted by $\pi^1\succeq\pi^2$) if and only if $f_i(\pi_1)\geq f_i(\pi_2)$ for all $i\in\{1,\cdots,m\}$ and $\mathbf{F}(\pi^1)\neq\mathbf{F}(\pi^2)$.
\end{definition}

\begin{definition}
    A policy $\pi^\ast$ is said to be \underline{Pareto-optimal} if and only if $\nexists \pi^\prime$ such that $\pi^\prime\succeq\pi^\ast$.
\end{definition}

\begin{definition}
    The set of all Pareto-optimal policies is called the \underline{Pareto-optimal set} (PS), i.e., $\mathcal{PS}=\{\pi^\ast|\nexists\pi^\prime \text{ such that } \pi^\prime\succeq\pi^\ast\}$ and their corresponding objective vectors form the \underline{Pareto-optimal front} (PF), i.e., $\mathcal{PF}=\{\mathbf{F}(\pi^\ast)|\pi^\ast\in PS\}$.
\end{definition}

\subsection{Related Works}
\label{sec:related_works}

This section begins with a brief overview of existing research on MORL from both single- and multi-policy perspectives. Subsequently, we highlight selected works on preference learning for MORL and inverse reinforcement learning (IRL). For a more comprehensive literature review, interested readers can refer to excellent survey papers such as~\cite{RoijersVWD13,RoijersW17,HayesRBKMRVZDHH22,RadulescuMRN20}.

\subsubsection{Single-policy MORL}
\label{sec:single_morl}

Single-policy MORL methods rely on \textit{a priori} preferences for different objectives to first convert a multi-objective problem into a single-objective one through aggregation. Subsequently, any off-the-shelf RL algorithm can be applied to obtain a single optimal policy. Aggregation can take various forms, such as linear scalarization~\cite{GaborKS98,MannorS01} or more complex approaches. For instance, in ~\cite{Rolf20}, Rolf proposed an exponential function with the sum of different objectives as the exponential term. In~\cite{AbdolmalekiHNS20}, a distributional method was introduced to first calculate a set of policies, each addressing a selected objective and represented as a distribution over actions. A new policy is then synthesized by combining all these policies. Although single-policy MORL algorithms are easy to implement, their practical applicability is limited due to the unrealistic assumption of exact a priori preferences.

\subsubsection{Multi-policy MORL}
\label{sec:multi_morl}

Multi-policy MORL methods decompose the original problem into several subproblems. In principle, any scalarization method can be employed to formulate the subproblems. Subsequently, multiple single-objective RL algorithms are applied to solve these subproblems. Some researchers have proposed iterative frameworks to address different subproblems sequentially~\cite{MossalamARW16,YangSN19,ReymondMNA19}. This approach involves sampling one subproblem at a time, optimizing the corresponding policy, and then switching to another subproblem based on the performance of the previous policy optimization. Alternatively, there have been attempts to solve all subproblems simultaneously within a parallel framework~\cite{ParisiPSBR14,XuTMRSM20,LiZW21}. Multi-policy MORL can discover a set of Pareto-optimal policies without any a priori information or adaptation during the policy optimization process. However, it incurs higher computational and storage costs compared to single-policy MORL, which may be detrimental when dealing with problems involving more than two objectives.

\subsubsection{IRL and Preference Learning}
\label{sec:preference_learning}

An emerging research direction involves incorporating human-in-the-loop reinforcement learning by proactively interacting with decision makers (DMs) and learning their preferences, known as preference learning~\cite{WirthANF17,PeschlZOS22}. This is especially useful and practical when the DM's exact demand is not available beforehand. Manually crafted reward functions notoriously struggle to align with the DM's true aspirations~\cite{AmodeiOSCSM16}. Preference-based reinforcement learning can be roughly divided into two categories: preference-based IRL and preference-based MORL.

Preference-based IRL does not assume a reward function before the algorithm; instead, it infers one from observed demonstrations according to the DM's preference. In works such as~\cite{SugiyamaMM12,PanS18,BrownGNN19}, the algorithms attempt to update the reward function to fit given preference relationships as closely as possible, using preferred demonstrations as input. While these approaches consider the DM's preferences, it can be challenging to explicitly define all preference relationships a priori~\cite{BrownGN19}. In~\cite{KollmitzKBB20}, a direct method is proposed that adds an offset to the policy's output during interaction with the DM. However, the policy optimization process remains unmodified, limiting the influence of preference information. \cite{WirthFN16,ChristianoLBMLA17} further learn a parameterized representation of the DM's preference interactively, guiding the learning process of reward functions and optimal policies.

Preference-based MORL, the second category, has gained popularity in recent years and is the focus of this paper. Unlike IRL, preference-based MORL addresses a well-defined MOMDP problem and learns a utility function representing the trade-off between different objective functions based on the DM's feedback. For example, \cite{RoijersW17,RoijersMZMLPNA18,WanigasekaraLGL19} formulate the preference learning problem as a multi-armed bandit problem, learning the utility function through the DM's feedback in the form of pairwise comparisons or ranked sequences. These methods provide good examples of modeling and iteratively learning the DM's preferences. However, continuous, high-dimensional RL problems lack the availability of comparisons between a set of fixed arms. In ~\cite{YangSN19}, Yang et al. proposes \textit{a posteriori} policy adaptation in continuous environments, approximating the PF and using a Gaussian process-based preference model to learn the DM's preference and guide preferred policy selection. However, many trade-off alternative policies on the PF may not align with the DM's preferences, making the search for the entire PF a waste of computational resources. Furthermore, trade-off alternative policies outside the region of interest can introduce irrelevant or noisy information during the decision-making process. In the sibling area of evolutionary multi-objective optimization and MCDM, our group has made a series of contributions towards preference-based methods in a priori~\cite{CaoKWL14,LiDAY17,LiCMY18}, a posterior~\cite{ZouJYZZL19,NieGL20,ChenL21b,LiNGY22}, and interactive~\cite{Li19,LiCSY19,LiLY23} manners. In particular, our previous study has empirically demonstrated the effectiveness of leveraging user preferences in the search of solutions of interest~\cite{LiLDMY20}.

%% file: method.tex

\section{Proposed Method}
\label{sec:method}


\begin{figure*}[t!]
    \centering
    \includegraphics[width=\linewidth]{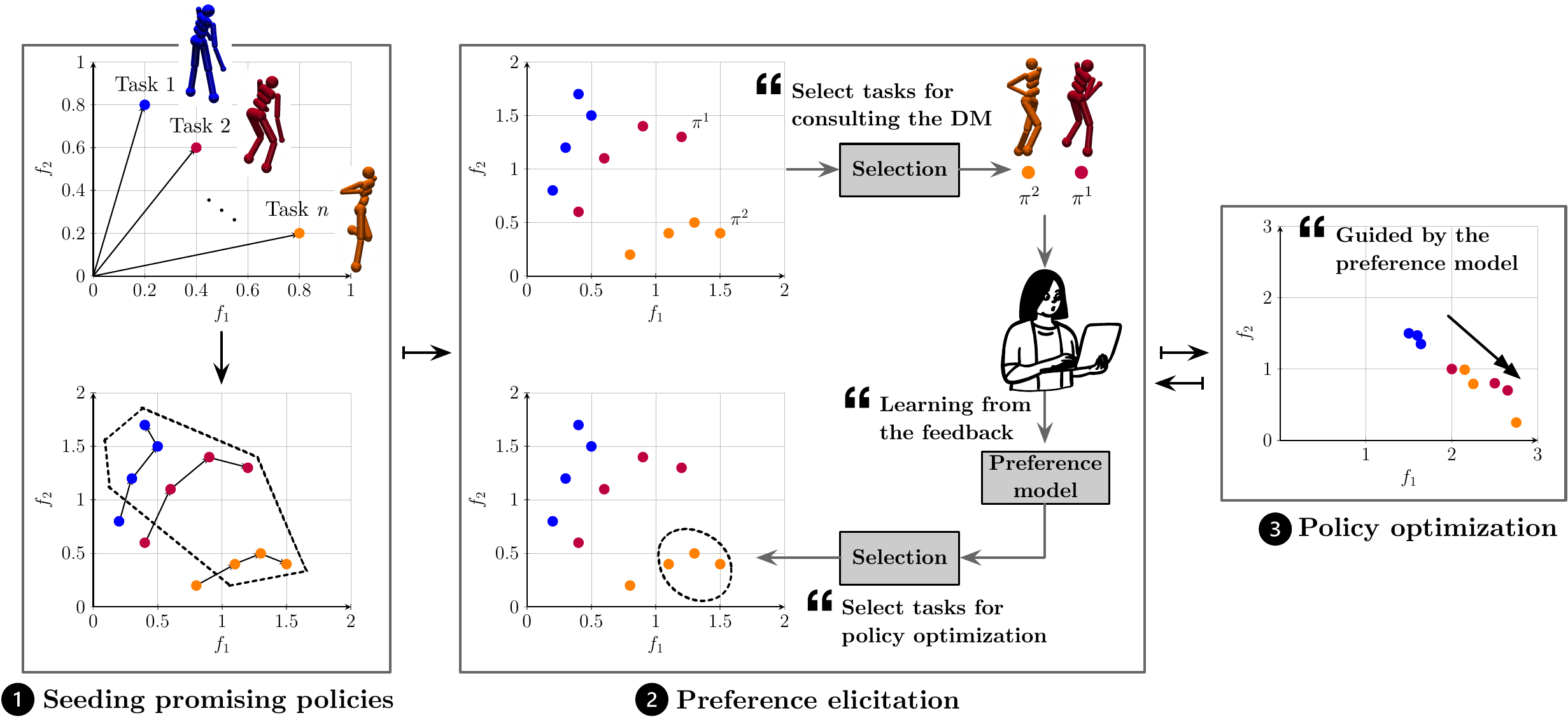}
    \caption{Flowchart of \our. It iterates between the \texttt{preference elicitation} and the \texttt{policy optimization} modules until the stopping criterion is met, and outputs the preferred policies.}
    \label{fig:flowchart}
\end{figure*}

\pref{fig:flowchart} presents the modular flow chart of our proposed \our\ framework. It consists of three core modules: $i$) the \texttt{seeding} module that obtains a set of promising seeding policies; $ii$) the \texttt{preference elicitation} module that plays as the interface where the DM interacts with the MORL and it learns the DM's preference from her feedback; and $iii$) the \texttt{policy optimization} module that leverages the learned preference to guide the search for the policy of interest. Note that the last two modules iterate between each other until the stopping criterion is met. In the following paragraphs, the implementation of each module is delineated step by step.

\subsection{Seeding Promising Policies}
\label{sec:seeding}

The \texttt{seeding} module operates as a conventional MORL, disregarding the DM's preference information and instead searching for a set of trade-off policies that approximate the PF. In this paper, our fundamental approach for MORL is inspired by the prevalent multi-objective evolutionary algorithm based on decomposition (MOEA/D)~\cite{ZhangL07,LiFKZ14,LiZKLW14,LiKZD15,WuKZLWL15,LiDZK15,LiDZZ17,WuLKZZ17,WuLKZ20}. We propose decomposing the MORL problem into a collection of subproblems, each representing a linear aggregation of different objective functions:
\begin{equation}
    \tilde{J}(\pi,\mathbf{w})=\sum_{i=1}^mw_iJ_{i}(\pi),
    \label{eq:ws}
\end{equation}
where $\mathbf{w}=(w_1,\ldots,w_m)^\top$ and $\sum_{i=1}^mw_i=1$ is the weight vector that represents certain preferences for different objectives. Note that any aggregation function can be applied in the \our\ framework, while we will investigate this aspect in~\pref{sec:rq6} of our empirical study. To balance convergence and diversity, we employ the Das and Dennis' method~\cite{DasD98,WuKJLZ17,WuLKZZ19,PruvostDLL020} to define a set of weight vectors $\mathcal{W}=\{\mathbf{w}^i\}_{i=1}^N$ evenly distributed along a unit simplex. Here, $N=$ ${H+m-1}\choose{m-1}$, and $H>1$ is the number of equally spaced divisions along each coordinate.

In the \our\ framework, all subproblems are handled simultaneously in a parallel manner, with each treated as an individual RL task. It is important to note that each RL policy is associated with a dedicated weight vector (i.e., a subproblem) in our \our\ framework. Given the continuous state and action spaces of the RL tasks considered in this paper, we employ proximal policy optimization (PPO)~\cite{SchulmanWDRK17}, a policy-gradient algorithm that optimizes the policy by explicitly updating the policy network $\pi_\theta$ using a gradient-based method. The objective function of the PPO is:
\begin{equation}
    \begin{aligned}
        J^{\mathrm{clip}}(\pi_\theta)=&\mathbb{E}_t\bigg[\min\bigg(\frac{\pi_{\theta}(a_t|s_t)}{\pi_{\theta_{\mathrm{old}}}(a_t|s_t)}\hat{A}_t,\\& \mathrm{clip}\left(\frac{\pi_{\theta}(a_t|s_t)}{\pi_{\theta_{\mathrm{old}}}(a_t|s_t)},1-\epsilon,1+\epsilon\right)\hat{A}_t\bigg)\bigg],
    \end{aligned}
    \label{eq:ppo}
\end{equation}
where $\hat{A}_t$ is an advantage function that estimates the performance of the current state-action tuple $\langle s_t,a_t\rangle$, and $\epsilon\in\left(0,1\right)$ controls the range of policy updates to avoid excessive optimization. The pseudo code of this multi-objective PPO (\texttt{MOPPO}) is given in~\pref{alg:moppo}, the output of which is the set of non-dominated policies, denoted as $\Pi$.

\begin{remark}
    The motivation of this \texttt{seeding} module stems from our preliminary experiments, which showed that interacting with the DM and conducting preference learning at the beginning of \our\ is ineffective. This observation can be attributed to the fact that the initial trade-off policies are far from the PF, which can be misleading for the DM during preference elicitation. As a result, we introduce the \texttt{seeding} module to first search for a set of reasonably good trade-off policies before proceeding with preference learning. This approach aims to provide a better starting point for the subsequent steps in the \our\ framework.
\end{remark}

\begin{algorithm}[t!]
    \caption{\texttt{MOPPO}}
    \label{alg:moppo}
    \KwIn{$t_\mathrm{r}$: \# of environment steps, $\overline{\Pi}$: stored policies}
  	Initialize $\Pi:=\emptyset$ for storing optimized policies\;
	\ForEach{$\langle\pi^i,\mathbf{w}^i\rangle\in\overline{\Pi}$}{\tcc{every task works parallelly}
        \While{the stopping criterion is not met}{
            Execute $\pi^i$ and store the collected trajectory data in a buffer $\mathcal{S}^i:=\left(\langle s^i_1,a^i_1,r^i_1\rangle,\ldots,\langle s^i_{t_\mathrm{r}},a^i_{t_\mathrm{r}},r^i_{t_\mathrm{r}}\rangle\right)$\;
            \While{the stopping criterion is not met}
            {
                Sample a trajectory from $\mathcal{S}^i$, and use $\mathbf{w}^i$ to aggregate its rewards\;
                Calculate $\mathcal{\hat{A}}_t$ as mentioned in~\cite{SchulmanMLJA15}\;
                Update $\pi^i$ by optimizing $J^{\mathrm{clip}}(\pi^i)$ in~\pref{eq:ppo}\;
            }
            Insert updated task $\langle\pi^i,\mathbf{w}^i\rangle$ to $\Pi$\;
        }
    }
    \Return Non-dominated policies in $\Pi$
\end{algorithm}

\subsection{Preference Elicitation}
\label{sec:preference_elicitation}

This module is the crux of proposed \our, whose pseudo code is given in~\pref{alg:pl}. It consists of three steps including \texttt{consultation}, \texttt{preference learning}, and \texttt{preference translation}. They are designed to address the following three questions. 

\subsubsection{How to collect the DM's feedback}
\label{sec:interaction}

In the \texttt{preference elicitation} module, the \texttt{consultation} step serves as an interface where the DM is asked to provide feedback on the quality of selected policies based on their preferences. In this work, we consider an indirect preference information in the form of holistic pairwise comparisons of policy pairs $\langle\pi^1,\pi^2\rangle$. The outcome of this pairwise comparison can be $\pi^1$ being \textit{better}, \textit{worse}, or \textit{indifferent} compared to $\pi^2$, denoted as $\pi^1\succ\pi^2$, $\pi^1\prec\pi^2$, or $\pi^1\simeq\pi^2$, respectively. Asking the DM to compare all ${|\Pi|}\choose{2}$ policy pairs derived from the seeding policies obtained in the \texttt{seeding} module may overwhelm them. To alleviate this issue, we only choose the two most informative policies from $\Pi$ to consult the DM. Our preliminary experiments showed that this approach not only significantly reduces the DM's workload but is also sufficient for preference learning in \our. Inspired by~\cite{Auer02}, we evaluate the information of a policy $\pi\in\Pi$ as:
\begin{equation}
    \mathtt{I}(\pi)=\mu(\pi)+\alpha\sqrt{{\sigma(\pi)\tilde{n}}/{\tilde{n}_\pi}},
    \label{eq:information}
\end{equation}
where $\mu(\pi)$ and $\sigma(\pi)$ represent the mean and variance predicted by the preference model, which will be introduced in~\pref{sec:preference_model}. $\tilde{n}_\pi$ denotes the number of times $\pi$ has been queried in previous interactions, $\tilde{n}$ represents the total number of queries incurred so far, and $\alpha>0$ is a parameter that balances exploitation and exploration. Based on this definition, two policies $\pi^1=\argmax_{\pi\in\Pi}\mathtt{I}(\pi)$ and $\pi^2=\argmax_{\pi\in\Pi\setminus{{\pi^1}}}\mathtt{I}(\pi)$ are chosen for the \texttt{consultation} step. The queried policies are stored in a dedicated set $\tilde{\Pi}$.

\begin{remark}
    In principle, we can employ \texttt{MOPPO}, as in the \texttt{seeding} module, to obtain a set of policies that approximate the entire PF. However, these non-dominated policies may be too diverse for DMs to identify their preferred ones that satisfy their aspirations in practice. Thus, a more refined approach is required to accommodate the DM's specific preferences.
\end{remark}

\begin{remark}

    At the beginning of the \texttt{preference elicitation} module, no data has been collected from the DM to train the preference model. To address this, during the first $\tau > 1$ iterations, we randomly select two policies in the \texttt{consultation} step. This approach allows us to gather initial data and subsequently refine the preference model as more information becomes available.
\end{remark}

\begin{remark}
    In the context of MORL, the DM's preference can be represented by assigning weights to different objective functions based on their importance~\cite{WrayZM15,NatarajanT05,YangSN19,WirthFN16}. However, this presents a paradox in our context, as the preference is unknown a priori. An alternative approach is to assign a numeric score to each candidate~\cite{LiCSY19}. Nevertheless, as discussed in~\cite{ZintgrafRLJN18}, this method can be vague and challenging for non-domain experts to comprehend and apply effectively.
\end{remark}

\begin{remark}
    Note that the \texttt{preference elicitation} module in \our\ is highly efficient where we only need consult the DM once at each round with a pair of candidates.
\end{remark}

\subsubsection{How to model the DM's preference information}
\label{sec:preference_model}

Based on the holistic pairwise comparison(s) collected in the \texttt{consultation} step, the goal of the \texttt{preference learning} step is to learn a preference model that evaluates the quality of solutions according to the DM’s preference. Here our preference model is represented as a utility function $u(\pi):\mathbb{R}^m\rightarrow\mathbb{R}$ such that the ranking order of a set of testing policies $\hat{\Pi}$ satisfy that $\forall\pi^1,\pi^2\in\hat{\Pi}$, $u(\pi^1)>u(\pi^2)$ if $\pi^1\succ\pi^2$ and $u(\pi^1)=u(\pi^2)$ if $\pi^1\simeq\pi^2$. This paper applies a zero-mean Gaussian process (GP)~\cite{GPML} with the radial basis function kernel to serve as the regression model. In particular, $\tilde{\Pi}=\{\tilde{\pi}^i\}_{i=1}^\kappa$ is used as the training data where $\kappa$ is the number of policies queried at the \texttt{consultation} step. Then, we train $u(\pi)$ based on maximum a posterior estimation as:
\begin{equation}
    u^\star=\argmax\limits_u\mathbb{P}(u)\mathbb{P}(\tilde{\Pi}|u),
    \label{eq:map}
\end{equation}
where $\mathbb{P}(u)$ is a zero-mean Gaussian and $\mathbb{P}(\tilde{\Pi}|u)=\prod\limits_{i=1}^{\frac{\kappa}{2}}\Phi(\frac{u(\pi_i^1)-u(\pi_i^2)}{\sqrt{2}})$ is 
the likelihood of $u(\pi)$, $\Phi(\cdot)$ is the standard normal cumulative distribution. As in~\cite{ChuG05}, for each pair of $\left\langle{u(\pi_i^1),u(\pi_i^2)}\right\rangle$ where $\pi_i^1\succ\pi_i^2$, solving~\pref{eq:map} is equivalent to solving the following minimization problem: 
\begin{equation}
    \mathrm{minimize} -\sum\limits_{i=1}^{\frac{\kappa}{2}}\bigg{[}\Phi(\frac{u(\pi_i^1)-u(\pi_i^2)}{\sqrt{2}})+\frac{1}{2}u^{\top}(\pi_i^1)K^{-1}u(\pi_i^2)\bigg{]},
    \label{eq:minimize}
\end{equation}
where $K$ is the covariance matrix of $\tilde{\Pi}$. By 
applying Newton-Raphson formula, we can finally get the numerical approximation of the real $u^\star$. Given a testing policy $\pi^\star$, the mean and variance of $u^\star(\pi^\star)$ are predicted as 
$\mathrm{\mu}(\pi^\star)={\mathbf{k}^\star}^{\top}K^{-1}\mathbf{U}$ and 
$\mathrm{\sigma}(\pi^\star)=k(\pi^\star,\pi^\star)-{\mathbf{k}^\star}^{\top}(K+\Lambda^{-1})^{-1}{\mathbf{k}^\star}$, respectively, 
where $\mathbf{U}=(u^\star(\tilde{\pi}^1),\ldots,u^\star(\tilde{\pi}^\kappa))^\top$. $\mathbf{k}^\star$ is the covariance vector between $\tilde{\Pi}$ and $\pi^\star$. $\Lambda^{-1}$ is a $\kappa\times\kappa$ matrix for estimating $\mathbf{U}$.

\begin{algorithm}[t!]
    \caption{\texttt{preference elicitation}}
    \label{alg:pl}
    \KwIn{$\Pi$: stored non-dominated policies, $\tilde{\Pi}$: queried policies as training data, $u(\pi)$: preference model}
    \tcc{consultation step}
    \eIf{random selection}{
        Randomly sample two policies $\pi^1$, $\pi^2$ from $\Pi$.
        }{
        \ForEach{$\pi\in\Pi$}{
        	Evaluate its utility value $u(\pi)$ and $\mathtt{I}(\pi)$\;}
        $\pi^1:=\argmax_{\pi\in\Pi}\mathtt{I}(\pi)$, $\pi^2:=\argmax_{\pi\in\Pi\setminus{\{\pi^1\}}}\mathtt{I}(\pi)$\;
        }
        Query the DM with $\pi^1$ and $\pi^2$, store the result in $\tilde{\Pi}$\;
        \tcc{preference learning step}
        Use the DM's feedback to update $\tilde{\Pi}$ and the preference model $u(\pi)$ as in~\pref{sec:preference_model}\;
        \tcc{preference translation step}
        Generate the new policy set $\Pi$ as in~\pref{sec:translation}\; 
        \Return $\Pi$, $\tilde{\Pi}$, $u(\pi)$.
\end{algorithm}

\subsubsection{How to leverage the learned preference information}
\label{sec:translation}

\begin{figure}[htbp]
    \centering
    \includegraphics[width=.85\linewidth]{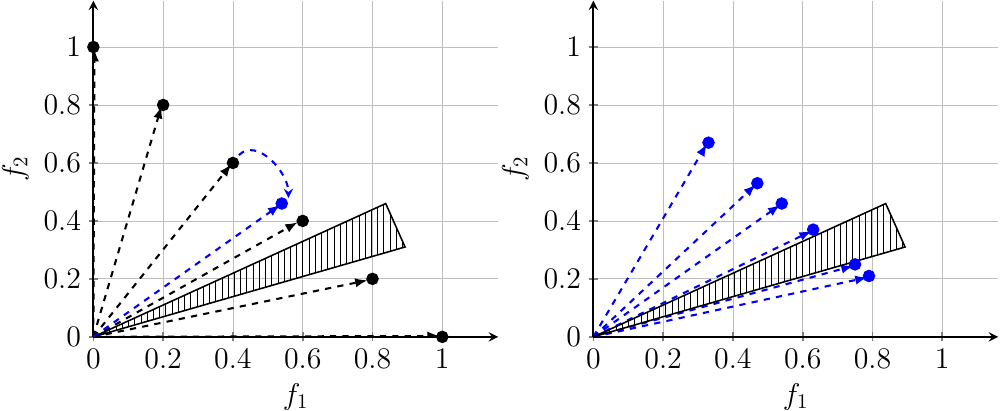}
    \caption{Illustration of the weight vector adjustment in Step 4. (a) The ROI is highlighted as the shaded cone region versus the evenly distributed weight vectors (denoted as $\newmoon$). (b) Adjusted weight vectors towards the ROI (denoted as \textcolor{blue}{$\newmoon$}).}
    \label{fig:preference_elicitation}
\end{figure}

In the \texttt{preference translation} step, the learned preference information is converted into a format that can be used to guide the \texttt{policy optimization} module. In the context of our proposed \our, this involves generating a new policy archive $\Pi$ for the next round of the policy optimization. This process consists of five steps:
\begin{enumerate}[Step 1:]
    \item Assign a preference score $\psi(\pi)=\mu(\pi)+\beta\sigma(\pi)$ to each policy $\pi\in\Pi$.
    \item Rank the policies in $\Pi$ based on the preference scores from Step 1. Identify the top $\kappa_1$ policies to construct a temporary policy archive $\dot{\Pi}$. Store the weight vectors associated with these policies, considered as promising, in a temporary archive $\dot{\mathcal{W}}=\{\dot{\mathbf{w}}^i\}_{i=1}^{\kappa_1}$.
    \item Generate a set of evenly distributed weight vectors $\tilde{\mathcal{W}}=\{\tilde{\mathbf{w}}^i\}_{i=1}^{\tilde{N}}$, as in the \texttt{seeding} module, where $\tilde{N}>N$. Set another temporary archive $\ddot{\mathcal{W}}=\emptyset$.
    \item Generate a set of weight vectors biased towards the region of interest (ROI), as illustrated in~\pref{fig:preference_elicitation}. Then, for $i=1$ to $\tilde{N}$, perform the following:
        \begin{enumerate}[Step 4.1:]
            \item Determine the reference point of $\tilde{\mathbf{w}}^i$ as $\mathbf{w}^\mathrm{r}=\argmin_{\dot{\mathbf{w}}\in\dot{\mathcal{W}}}|\dot{\mathbf{w}}-\tilde{\mathbf{w}}^i|$.
            \item Move $\tilde{\mathbf{w}}^i$ towards $\mathbf{w}^\mathrm{r}$ as $\ddot{w}^i_j=\tilde{w}^i_j+\eta(w^\mathrm{r}_j-\tilde{w}^i_j)$, where $j\in\{1,\cdots,m\}$.
            \item Update $\ddot{\mathcal{W}}$ with the new weight vector: $\ddot{\mathcal{W}}=\ddot{\mathcal{W}}\bigcup\{\ddot{\mathbf{w}}\}$ and return to Step 4.
        \end{enumerate}
    \item Randomly select $\kappa_2$ weight vectors from $\ddot{\mathcal{W}}$ and remove the others to form a reduced $\ddot{\mathcal{W}}$. Randomly choose $\kappa_2$ policies from $\dot{\Pi}$ and associate each selected policy with a weight vector in $\ddot{\mathcal{W}}$ to form a temporary archive $\ddot{\Pi}$. Construct the new policy archive $\Pi$ for the next round of the policy optimization as $\Pi=\dot{\Pi}\bigcup\ddot{\Pi}$.
\end{enumerate}

\begin{remark}
    In Step 1, $\psi(\pi)$ serves as an approximate value function that evaluates the satisfaction of policy $\pi$ concerning the DM's preference information. Notably, the parameter $\beta>0$ helps strike a balance between exploitation and exploration in the \texttt{PBMORL} process.
\end{remark}

\begin{remark}
    Note that the preference score of a policy is distinct from the information of a policy, as defined in~\pref{eq:information}. This difference arises because the objective of Step 2 is to identify policies that best represent the DM's implicit preferences. By making this distinction, our approach effectively captures the nuances of preference elicitation.
\end{remark}

\begin{remark}
    In our preliminary experiments, we observed that MORL can become overly exploitative and prone to being trapped in local optima when the \texttt{policy optimization} module is solely driven by $\dot{\mathcal{W}}$. To address this issue, Steps 3 and 4 are introduced to incorporate additional exploration outside of the ROI. This approach effectively balances exploitation and exploration, ensuring a more robust optimization process
\end{remark}

\begin{remark}
    In Step 5, the composition of $\mathcal{W}$ is designed to strike a balance between exploitation and exploration, with a focus on enhancing exploitation to accelerate convergence. In practice, we set $\kappa_1=80\%\times|\Pi|$ and $\kappa_2=20\%\times|\Pi|$. We will conduct a parameter sensitivity study on the settings of $\kappa_1$ and $\kappa_2$ in~\pref{sec:rq4} of our empirical study.
\end{remark}

\vspace{-1em}
\subsection{Policy Optimization}
\label{sec:policy_optimization}

In the \texttt{policy optimization} module, we employ the \texttt{MOPPO} algorithm, as done in the \texttt{seeding} module introduced in~\pref{sec:seeding}, which utilizes the biased weight vectors generated in the \texttt{preference elicitation} module to guide the MORL process. It is important to note that the outputs of this \texttt{policy optimization} module are fed back into the \texttt{preference elicitation} module, further enhancing the navigation towards the DM's preferred non-dominated policies until the stopping criterion are met.

\subsection{Time Complexity Analysis}
\label{sec:time complexity}

Since the \texttt{seeding} and the \texttt{policy optimization} modules of \our\ are MORL based on PPO, the corresponding time complexity is bounded by the PPO itself. Therefore, here we focus on analyzing the time complexity of the \texttt{preference elicitation} module, which consists of three steps. Specifically, during the \texttt{consultation} step, the preference model first evaluate the quality of policies in $\Pi$ which incurs $\mathcal{O}(|\Pi|)$ evaluations. Then, it chooses two elite policies to query the DM. After the DM's feedback is collected, the complexity of the \texttt{preference learning} step is bounded by $\mathcal{O}(\kappa^3)$, where $\kappa$ is the number of data instances in the training set. During the \texttt{preference translation} step, the computational complexity is mainly dominated by the ranking of policies in $\Pi$, i.e., $\mathcal{O}(|\Pi|\log |\Pi|)$. All in all, the time complexity of the \texttt{preference elicitation} module is $\max\left\{\mathcal{O}(\kappa^3),\mathcal{O}(|\Pi|\log |\Pi|)\right\}$.

%% file: settings.tex

\section{Experimental Settings}
\label{sec:settings}

This section introduces the settings of our empirical study including the benchmark problems, the peer algorithms, and the performance metrics.

\subsection{Benchmark Problems}
\label{sec: benchmark problems}

In empirical study, we consider two types of benchmark problems: one is from the popular \textsc{MuJoCo} environment~\cite{TodorovET12} and the other is a multi-microgrid system design (MMSD) problem~\cite{ChiuSP15}.

\subsubsection{\textsc{MuJoCo} environment}
\label{sec:mujoco}

Following~\cite{XuTMRSM20}, we examine seven benchmark problems developed from \textsc{MuJoCo}. Their objective functions and search spaces are outlined below:
\begin{itemize}
    \item \texttt{Ant-v2}: This problem uses the speeds at the $x$ and $y$ axes as the two objectives. The state space is $\mathcal{S}\in{\mathbb{R}^{27}}$, and the action space is $\mathcal{A}\in{\mathbb{R}^8}$.

    \item \texttt{HalfCheetah-v2}: Two objectives are considered, including forward speed and energy consumption. The state space is $\mathcal{S}\in{\mathbb{R}^{17}}$, and the action space is $\mathcal{A}\in{\mathbb{R}^6}$.

    \item \texttt{Hopper-v2}: This problem considers forward speed and jumping height as the objectives. The state space is $\mathcal{S}\in{\mathbb{R}^{11}}$, and the action space is $\mathcal{A}\in{\mathbb{R}^3}$.

    \item \texttt{Humanoid-v2}: Two objectives are evaluated, including forward speed and energy consumption. The state space is $\mathcal{S}\in{\mathbb{R}^{376}}$, and the action space is $\mathcal{A}\in{\mathbb{R}^{17}}$.

    \item \texttt{Swimmer-v2}: This problem focuses on forward speed and energy consumption objectives. The state space is $\mathcal{S}\in{\mathbb{R}^8}$, and the action space is $\mathcal{A}\in{\mathbb{R}^2}$.

    \item \texttt{Walker2d-v2}: Two objectives are considered, including forward speed and energy consumption. The state space is $\mathcal{S}\in{\mathbb{R}^{17}}$, and the action space is $\mathcal{A}\in{\mathbb{R}^6}$.

    \item \texttt{Hopper-v3}: This problem incorporates three objectives, including forward speed, jumping height, and energy consumption. The state space is $\mathcal{S}\in{\mathbb{R}^{11}}$, and the action space is $\mathcal{A}\in{\mathbb{R}^3}$.
\end{itemize}

\subsubsection{\textsc{MMSD} environment}
\label{sec:mmsd}

\begin{table*}[t!]
    \centering
    \caption{Lookup table of the mathematical notations used in the \textsc{MMSD} problem~\cite{ChiuSP15}.}
    \label{tab:mmsd}
    \resizebox{.9\linewidth}{!}{ 
        \begin{tabular}{c|l}
            \toprule
            \textsc{Symbol} & \textsc{Description} \\
            \midrule
            $n$ & The number of microgrids\\
            $n_\mathrm{s}$ & The number of microgrids with power storage\\
            $t$    &  The current time step \\
            $T$&The length of one episode\\
            $P_g(t)$&The sum of power given to all of microgirds at the $t$-th time step\\
            $p^g_i(t)$&The power given to the $i$-th microgrid at $t$-th time step\\
            $p^d_i(t)$&The power demand of the $i$-th microgrid at the $t$-th time step\\ 
            $\lambda{(t)}$&The power price at the $t$-th time step\\
            $s_i(t)$&The power storage of the $i$-th microgrid at the $t$-th time step\\
            $v_i(t)$& The energy that the distributed power gives to the $i$-th microgrid at the $t$-th time step\\
            $b_i(t)$&The base load of the $i$-th microgrid at the $t$-th time step\\
            $c_i$&The maximum rate of storage charging and discharging\\
            $l_i$&The lower bound of the energy storage of the $i$-th microgrid\\
            $u_i$&The upper bound of the energy storage of the $i$-th microgrid\\
            \bottomrule
         \end{tabular}
    }
\end{table*}

It considers the following three-objective optimization problem:
\begin{equation}
    \begin{array}{l l}
        \mathrm{maximize}\;\;\; \mathbf{r}=\left(\mathrm{U}_\mathrm{pg}(T), \mathrm{U}_\mathrm{mg}(T), \sum_{i=1}^{n_{s}}s_{i}(T)\right)^\top\\
        \mathrm{subject\ to} \;\; \|s_i(t)-s_i(t-1)\|<c_i, \forall i\in\{1,\ldots,n_\mathrm{s}\}
    \end{array},
    \label{eq:mmsd}
\end{equation}
where $\mathbf{r}\subseteq\mathbb{R}^3$ is the reward, $l_i<s_i(t)<u_i$. The mathematical definitions of these three objectives are delineated as follows, while the related notations are listed in~\pref{tab:mmsd}.
\begin{itemize}
    \item $\mathrm{U}_\mathrm{pg}(T)$ evaluates the utility value achieved by the power grid after one episode:
\begin{equation}
\mathrm{U}_\mathrm{pg}(T)=\sum_{t=1}^T\sum_{i=1}^n\left(\mathrm{U}(p^d_i(t),w_{i})-\lambda(t)p^d_i(t)\right),
\end{equation}
where
\begin{equation}
    \mathrm{U}(p_i^d(t),w_{i})=
        \begin{dcases}
            \frac{w_{i}}{\alpha}, &\text{if } p_i^d(t)\geq\frac{w_{i}}{\alpha} \\
            w_ip^d_i(t)-\frac{\alpha{p^d_i}(t)^2}{2}, &\text{if } 0\leq p^d_i(t)\leq\frac{w_{i}}{\alpha}
        \end{dcases},
\end{equation}
where $w_{i}$ and $\alpha$ are pre-defined parameters.
    \item $\mathrm{U}_\mathrm{mg}(T)$ is the total utility value achieved by all microgrids after one episode:
\begin{equation}
    \mathrm{U}_\mathrm{mg}(T)=\sum_{t=1}^T\lambda(t)P_g(t)-\beta P_g(t)^{2},
\end{equation}
where $\beta$ is a pre-defined parameter.
    \item To measure the stability of a multi-microgrid system, we use $\sum_{i=1}^{n_{s}}s_{i}(T)$ to evaluate its total energy storage. In particular, we have:
\begin{equation}
    s_i(t)=s_i(t-1)+p^g_i(t)-p^d_i(t)+v_i(t),
\end{equation}
where $p^d_i(t)$ is decided by both the base load of the $i$-th microgrid~\cite{ChiuSP15} and a scaling factor:
\begin{equation}
    p^d_i(t)=(1+h_i(t))b_i(t),
\end{equation}
where $h_i(t)$ is defined as:
\begin{equation}
    h_i(t)=
        \begin{dcases}
            0.01\lambda^2(t)-0.12\lambda(t)+0.26, &\text{if } i=1 \\
            -0.01\lambda^2(t)+0.13, &\text{if } i=2 \\
            -0.01\lambda^2(t)+0.02\lambda(t)+0.08, &\text{if } i=3
        \end{dcases}.
\end{equation}
The state space is $\mathcal{S}\subseteq\mathbb{R}^{n+2}$ where $\forall\mathbf{s}\in\mathcal{S}$, we have $\mathbf{s}=\left(t,p^g_1(t),\ldots,p^g_n(t),\lambda(t)\right)^\top$. The action space is $\mathcal{A}\subseteq\mathbb{R}^{n_\mathrm{s}+1}$ where $\forall a\in\mathcal{A}$, we have $a=\left(\Delta\lambda(t),\Delta{p^g_1(t)},\ldots,\Delta{p^g_{n_\mathrm{s}}(t)}\right)^\top$. 
\end{itemize}

\subsection{Peer Algorithms}
\label{sec:peer_algorithms}

In our experiments, we consider two types of peer algorithms. The first category comprises conventional MORL algorithms that do not take the DM's preference information into account.
\begin{itemize}
    \item \texttt{RA}~\cite{ParisiPSBR14}: \texttt{RA} transforms a MORL problem into several single-objective RL tasks by weighted aggregations. Different from our proposed \our, \texttt{RA} solves these tasks independently.

    \item \texttt{PGMORL}~\cite{XuTMRSM20}: Designed to optimize multiple policies in parallel, \texttt{PGMORL} iteratively adjusts the directions based on a predictive method.

    \item \texttt{MOIA}~\cite{ChiuSP15}: \texttt{MOIA} is a dedicated multi-objective evolutionary algorithm tailored for the multi-microgrid system design problem.
\end{itemize}

The second category consists of preference-based MORL algorithms from the literature:
\begin{itemize}
    \item \texttt{MORL-Adaptation}~\cite{YangSN19}: By generalizing the Bellman operator to MORL problems, \texttt{MORL-Adaptation} learns a universal parametric representation for all latent preferences. After the training phase, the learned policy can adapt to a given preference without further adaptation.

    \item \texttt{META-MORL}~\cite{ChenGBJ19}: \texttt{META-MORL} formulates MORL as a meta-learning problem conditioned on a task distribution over preferences. After the training phase, the learned policy can be adapted to a DM-specified preference through a fine-tuning phase.

    \item \texttt{MOMPO}~\cite{AbdolmalekiHNS20}: \texttt{MOMPO} learns an action distribution for each objective in each round of policy training and updates the policy by fitting it to a combination of action distributions. Before the training phase, a set of parameters representing the DM's preference for each objective is provided, controlling the learning rate of the corresponding action distribution.

    \item \texttt{MORAL}~\cite{PeschlZOS22}: \texttt{MORAL} first learns a multi-objective reward function from demonstrations. Then, similar to our proposed \our, it learns the DM's preference as a weight vector based on the Bradley-Terry model~\cite{BradleyT52} to guide the policy optimization.
\end{itemize}

All peer algorithms used the same number of environment steps, and \pref{tab:parameters} lists the key hyperparameters used in our experiments.

\begin{table*}[t!]
    \centering
    \caption{List of the hyperparameter settings used in our experiments.}
    \resizebox{.9\linewidth}{!}{
        \begin{tabular}{l|c|c|c}
            \toprule
            \textsc{Hyperparameter} & \textsc{MuJoCo} ($m=2$) & \textsc{MuJoCo} ($m=3$) & Microgrid\\
            \midrule
            \# of environment steps & $8\times{10^6}$ & $8\times{10^6}$ & $4\times{10^6}$ \\
            \# of environment steps for the \texttt{seeding} stage & $4\times{10^5}$ & $4\times{10^5}$ & $4\times{10^4}$ \\
            \# of interactions with the DM & $40$ & $40$ & $40$ \\
            \# of subtasks built before starting \our & $6$ & $21$ & $21$ \\
            \# of microgrids & \diagbox{}{} & \diagbox{}{} & $3$ \\
            \# of microgrids with energy storage & \diagbox{}{} & \diagbox{}{} & $2$ \\
            \bottomrule
         \end{tabular}
    }
    \label{tab:parameters}
\end{table*}

\subsection{Performance Metrics}
\label{sec:metrics}

As discussed in~\cite{LiDY18,LaiLL21,TanabeL23}, quality assessment of non-dominated trade-off policies is far from trivial when considering DM's preference information regarding multiple conflicting criteria. In our experiments, we consider the following two metrics to serve this purpose.
\begin{itemize}
    \item The first one is the \textit{approximation accuracy} that evaluates the closeness of the best non-dominated policy to the DM preferred one:
        \begin{equation}
            \epsilon^\star(\Pi)=\min_{\pi\in\Pi}\|\pi-\pi^\star\|_2,
        \end{equation}
        where $\|\cdot\|_2$ is the Euclidean norm, $\pi^\star$ is the golden policy specified by the DM but is unknown to the algorithm, and $\Pi$ is the non-dominated policy set obtained by MORL.

    \item The second one is the \textit{average accuracy} that evaluates the average distance of all non-dominated policies in $\Pi$ to the DM preferred one:
        \begin{equation}
            \bar{\epsilon}(\Pi)=\frac{\sum_{\pi\in\Pi}\|\pi-\pi^\star\|_2}{|\Pi|},
        \end{equation}
        where $|\cdot|$ is the cardinality of a set.
\end{itemize}

Note that calculating both $\epsilon^\star(\Pi)$ and $\overline{\epsilon}(\Pi)$ needs to specify a golden policy $\pi^\star$, which is represented as a $m$-dimensional hyperplane. Specifically, if we consider that the DM always prefers one objective over the others, $\pi^\star$ is defined as $f_i(\pi^\star)=10,000$, where $i$ is the objective index preferred by the DM. As for the \textsc{MMSD} problem, we set $f_1(\pi^\star)=0$ or $f_2(\pi^\star)=0$ if the DM prefers the first or the second objective, while we set $f_3(\pi^\star)=450$ if the DM prefers the third objective.

%% file: experiment.tex

\section{Experimental Results}
\label{sec:results}

In this section, we present and analyze the results obtained in our experiments.

\subsection{Comparison with Conventional MORL}
\label{sec:no_preference}

\begin{figure*}[t!]
	\center
    \includegraphics[width=\linewidth]{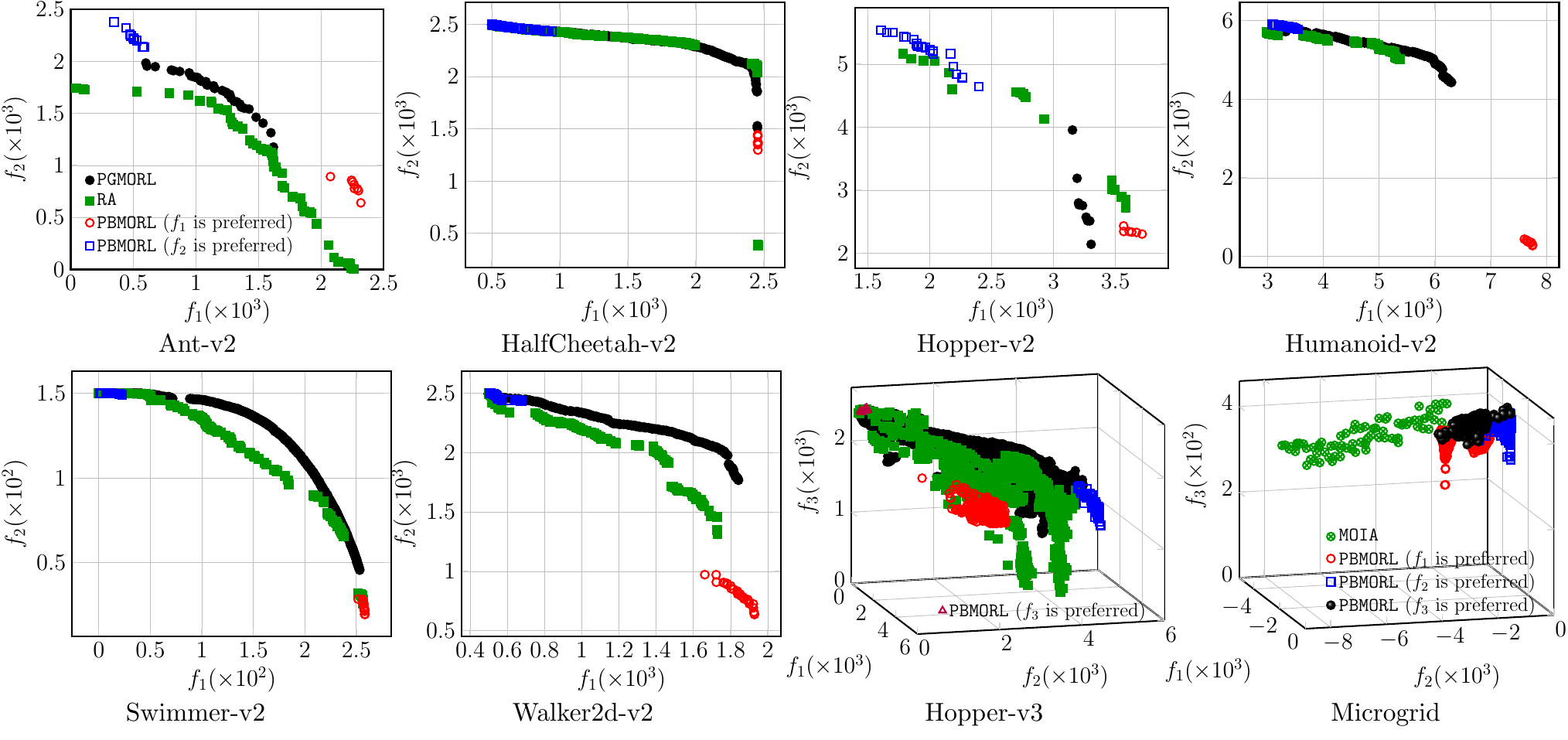}
    \caption{Plots of non-dominated policies obtained by \our\ versus \texttt{PGMORL}, \texttt{RA}, and \texttt{MOIA} with different preferences.}
    \label{fig:no_preference} 
\end{figure*}

\begin{table*}[t!]
    \centering
    \caption{Comparison results of $\epsilon^\star(\Pi)$ and $\overline{\epsilon}(\Pi)$ of \our\ versus \texttt{PGMORL}, \texttt{RA} and \texttt{MOIA} over $10$ runs with mean and standard deviation.}
    \resizebox{\linewidth}{!}{ 
    \begin{tabular}{c|c|cccccccc}
        \toprule    \multicolumn{1}{c}{\multirow{2}[4]{*}{}} &       & \multicolumn{2}{c|}{\our} & \multicolumn{2}{c|}{\texttt{PGMORL}} & \multicolumn{2}{c|}{\texttt{RA}} & \multicolumn{2}{c}{\texttt{MOIA}} \\
        \cmidrule{3-10}    \multicolumn{1}{c}{} & & $\epsilon^\star(\Pi)$ & \multicolumn{1}{c|}{$\overline{\epsilon}(\Pi)$} & $\epsilon^\star(\Pi)$ & \multicolumn{1}{c|}{$\overline{\epsilon}(\Pi)$} & $\epsilon^\star(\Pi)$ & \multicolumn{1}{c|}{$\overline{\epsilon}(\Pi)$} & $\epsilon^\star(\Pi)$ & $\overline{\epsilon}(\Pi)$ \\
        \midrule
        \multirow{1}[4]{*}{\texttt{Ant-v2}} & $f_1$    & \cellcolor[rgb]{ .702,  .702,  .702}{$\mathbf{7.709(1.68E-3)}$} &\cellcolor[rgb]{ .702,  .702,  .702}{$\mathbf{7.880(3.32E-2)}$} & $8.383$($5.07$E$-3$)&$8.898$($7.39$E$-2$)   & $7.795$($2.40$E$-3$)  & $8.506$($6.13$E$-3$)& \multicolumn{2}{c}{\multirow{15}[0]{*}{\diagbox{}}} \\
        \cmidrule{2-8}          & $f_2$    & \cellcolor[rgb]{ .702,  .702,  .702}{$\mathbf{7.637(6.98E-4)}$} &\cellcolor[rgb]{ .702,  .702,  .702}{$\mathbf{7.751(1.65E-4)}$} & $8.042$($9.23$E$-3$)  & $8.322$($1.57$E$-2$)  & $8.287$($6.99$E$-3$) & $8.965$($1.66$E$-3$)& \multicolumn{2}{c}{} \\
        \cmidrule{1-8}
        \multirow{1}[4]{*}{\texttt{HalfCheetah-v2}} & $f_1$    & \cellcolor[rgb]{ .702,  .702,  .702}{$\mathbf{7.541(3.21E-6)}$} &\cellcolor[rgb]{ .702,  .702,  .702}{$\mathbf{7.542(7.53E-6)}$} & $7.554$($1.84$E$-5$)& $8.584$($7.46$E$-4$)  & $7.542$($1.14$E$-5$)  & $8.583$($4.29$E$-4$)& \multicolumn{2}{c}{} \\
        \cmidrule{2-8}          & $f_2$    & \cellcolor[rgb]{ .702,  .702,  .702}{$\mathbf{7.500(3.70E-9)}$} & \cellcolor[rgb]{ .702,  .702,  .702}{$\mathbf{7.508(1.17E-4)}$} & $7.506$($2.31$E$-5$)& $7.729$($2.61$E$-4$)  & $7.504$($2.44$E$-5$)&$7.670$($4.61$E$-5$) & \multicolumn{2}{c}{} \\
        \cmidrule{1-8}
        \multirow{1}[4]{*}{\texttt{Hopper-v2}} & $f_1$    & \cellcolor[rgb]{ .702,  .702,  .702}{$\mathbf{6.221(1.14E-2)}$} & \cellcolor[rgb]{ .702,  .702,  .702}{$\mathbf{6.299(1.57E-2)}$} &  $6.744$($9.76$E$-4$)  & $6.891$($1.25$E$-1$)  & $6.467$($3.78$E$-2$)& $7.270$($9.95$E$-2$)& \multicolumn{2}{c}{} \\
        \cmidrule{2-8}          & $f_2$    & \cellcolor[rgb]{ .702,  .702,  .702}{$\mathbf{4.587(2.25E-2)}$} &\cellcolor[rgb]{ .702,  .702,  .702}{$\mathbf{4.972(3.11E-2)}$} & $6.160$($8.05$E$-2$)  & $7.305$($6.48$E$-2$)& $4.903$($3.99$E$-2$)&$5.910$($1.48$E$-3$) & \multicolumn{2}{c}{} \\
        \cmidrule{1-8}
        \multirow{1}[4]{*}{\texttt{Humanoid-v2}} & $f_1$    & \cellcolor[rgb]{ .702,  .702,  .702}{$\mathbf{2.362(2.51E-2)}$} &\cellcolor[rgb]{ .702,  .702,  .702}{$\mathbf{2.583(1.49E-1)}$} & $3.796$($1.19$E$-2$)& $5.066$($2.95$E$-2$)  & $4.719$($9.07$E$-2$)  & $5.783$($1.80$E$-2$)& \multicolumn{2}{c}{} \\
        \cmidrule{2-8}          & $f_2$    & \cellcolor[rgb]{ .702,  .702,  .702}{$\mathbf{4.099(6.96E-7)}$} & \cellcolor[rgb]{ .702,  .702,  .702}{$\mathbf{4.143(9.79E-5)}$} & $4.393$($2.76$E$-2$)  & $5.06$($6.93$E$-3$)  & $4.451$($2.06$E$-2$)& $4.67$($1.54$E$-3$)& \multicolumn{2}{c}{} \\
        \cmidrule{1-8}
        \multirow{1}[4]{*}{\texttt{Swimmer-v2}} & $f_1$    & \cellcolor[rgb]{ .702,  .702,  .702}{$\mathbf{9.747(1.12E-4)}$} & \cellcolor[rgb]{ .702,  .702,  .702}{$\mathbf{9.749(1.49E-4)}$} & $9.753$($1.25$E$-5$)&$9.842$($9.00$E$-5$)  & $9.756$($4.32$E$-4$)  & $9.913$($4.84$E$-4$)  & \multicolumn{2}{c}{}\\
        \cmidrule{2-8}          & $f_2$    & \cellcolor[rgb]{ .702,  .702,  .702}{$\mathbf{9.850(1.00E-12)}$}  &\cellcolor[rgb]{ .702,  .702,  .702}{$\mathbf{9.850(3.67E-12)}$} & \cellcolor[rgb]{ .702,  .702,  .702}{$\mathbf{9.850(1.08E-7)}$}  & $9.894$($1.80$E$-6$)  & $9.852$($1.11$E$-6$) & $9.898$($2.63$E$-5$) & \multicolumn{2}{c}{} \\
        \cmidrule{1-8}
        \multirow{1}[4]{*}{\texttt{Walker2d-v2}} & $f_1$    & \cellcolor[rgb]{ .702,  .702,  .702}{$\mathbf{8.048(9.77E-4)}$} & \cellcolor[rgb]{ .702,  .702,  .702}{$\mathbf{8.116(3.95E-3)}$} &  $8.189$($1.10$E$-2$) & $8.870$($2.90$E$-2$)  & $8.297$($5.48$E$-3$)  & $9.110$($1.23$E$-2$)& \multicolumn{2}{c}{} \\
        \cmidrule{2-8}          & $f_2$    & \cellcolor[rgb]{ .702,  .702,  .702}{$\mathbf{7.500(3.21E-8)}$}  &\cellcolor[rgb]{ .702,  .702,  .702}{$\mathbf{7.506(8.84E-5)}$} & \cellcolor[rgb]{ .702,  .702,  .702}{$\mathbf{7.500(1.23E-7)}$}& $7.786$($3.84$E$-4$)  & $7.502$($8.30$E$-6$) & $7.835$($2.54$E$-3$)  & \multicolumn{2}{c}{} \\
        \cmidrule{1-8}
        \multirow{2}[6]{*}{\texttt{Hopper-v3}} & $f_1$    & \cellcolor[rgb]{ .702,  .702,  .702}{$\mathbf{6.020(8.91E-8)}$}  &\cellcolor[rgb]{ .702,  .702,  .702}{$\mathbf{6.203(1.44E-3)}$} & $6.169$($1.74$E$-3$) & $7.56$($4.22$E$-3$)  & $6.283$($2.37$E$-3$) & $8.331$($1.55$E$-2$) & \multicolumn{2}{c}{} \\
        \cmidrule{2-8}          & $f_2$    & \cellcolor[rgb]{ .702,  .702,  .702}{$\mathbf{4.346(2.02E-3)}$}& \cellcolor[rgb]{ .702,  .702,  .702}{$\mathbf{4.539(1.97E-2)}$} &  $4.825$($1.89$E$-3$) & $7.131$($3.34$E$-3$) & $5.017$($4.73$E$-3$)     & $8.428$($2.40$E$-2$)  & \multicolumn{2}{c}{} \\
        \cmidrule{2-8}          & $f_3$    & \cellcolor[rgb]{ .702,  .702,  .702}{$\mathbf{7.500(1.90E-8)}$}  &\cellcolor[rgb]{ .702,  .702,  .702}{$\mathbf{7.502(1.23E-5)}$} & \cellcolor[rgb]{ .702,  .702,  .702}{$\mathbf{7.500(7.84E-8)}$} & $8.201$($4.15$E$-3$)  & $7.501$($6.27$E$-7$)&$7.984$($3.80$E$-4$)     & \multicolumn{2}{c}{}\\
        \cmidrule{1-10}
        \multirow{2}[6]{*}{\texttt{Microgrid}} & $f_1$    & \cellcolor[rgb]{ .702,  .702,  .702}{$\mathbf{1.149(8.85E-4)}$}&\cellcolor[rgb]{ .702,  .702,  .702}{$\mathbf{1.846(3.88E-5)}$} &\multicolumn{4}{c}{\multirow{3}[0]{*}{\diagbox{}}} & $1.845$($1.81$E$-3$)& $3.902$($2.56$E$-3$)  \\
        \cmidrule{2-4}\cmidrule{9-10}          & $f_2$    & \cellcolor[rgb]{ .702,  .702,  .702}{$\mathbf{0.000(4.87E-7)}$}&\cellcolor[rgb]{ .702,  .702,  .702}{$\mathbf{0.511(3.41E-6)}$} &\multicolumn{4}{c}{}   & $1.001$($7.35$E$-7$)  & $5.051$($4.09$E$-7$) \\
        \cmidrule{2-4}\cmidrule{9-10}          & $f_3$    &\cellcolor[rgb]{ .702,  .702,  .702}{$\mathbf{0.030(9.78E-8)}$}  &\cellcolor[rgb]{ .702,  .702,  .702}{$\mathbf{0.050(5.74E-8)}$} & \multicolumn{4}{c}{}        & $0.061$($2.14$E$-7$)  & $0.090$($5.22$E$-8$) \\
        \bottomrule
    \end{tabular}
    }
    \label{tab:no_preference}
\end{table*}

We first compare our proposed \our\ against three conventional MORL algorithms, which do not consider preferences. Note that since \texttt{RA} and \texttt{PGMORL} were merely designed for the \textsc{MuJoCo} environment while \texttt{MOIA} was deliberately designed for the \textsc{MMSD} environment, we only compare with them on their dedicated environment, respectively. To have a better visual interpretation, we plot the non-dominated policies found by different algorithms in~\pref{fig:no_preference}. Overall, the comparison results can be classified into two categorizes. The first category is on \texttt{Ant-v2}, \texttt{Swimmer-v2}, \texttt{Walker2d-v2}, and \texttt{Hopper-v3}. The non-dominated policies obtained by \texttt{RA} and \texttt{PGMORL} approximate a wide range of objective space, while the policies found by our proposed \our\ are notably biased towards the DM's specified objectives, i.e., the ROI. This explains the comparable results of \texttt{PBMORL} regarding \texttt{RA} and \texttt{PGMORL} on $\epsilon^\star(\Pi)$ shown in~\pref{tab:no_preference} where \our\ and \texttt{PGMORL} even achieve the same mean $\epsilon^\star(\Pi)$ values. However, if we refer to the $\overline{\epsilon}(\Pi)$, we can see that \our\ is significantly better, indicating that the average performance of \our\ in approximating the policies of interest outperforms \texttt{RA} and \texttt{PGMORL}. This is expected, as both \texttt{RA} and \texttt{PGMORL} identify too many policies outside the ROI. As for the other cases, i.e., \texttt{HalfCheetah-v2}, \texttt{Hopper-v2}, and \texttt{Humanoid-v2}, neither \texttt{RA} nor \texttt{PGMORL} can find the complete PF. In contrast, \our\ can discover reasonable non-dominated policies that meet the DM's preferred objectives. Consequently, as the comparison results of $\epsilon^\star(\Pi)$ and $\overline{\epsilon}(\Pi)$ shown in~\pref{tab:no_preference}, we can see that \our\ are consistently the best. As for the \textsc{MMSD} environment shown in~\pref{fig:no_preference}, we can see that the policies obtained by \texttt{MOIA} are completely dominated by those found by \our. This observation is also supported by the superior performance of $\epsilon^\star(\Pi)$ and $\overline{\epsilon}(\Pi)$ shown in~\pref{tab:no_preference}.

\subsection{Comparison with Preference-based MORL}
\label{sec:with_preference}

\begin{figure*}[t!]
	\center
    \includegraphics[width=\linewidth]{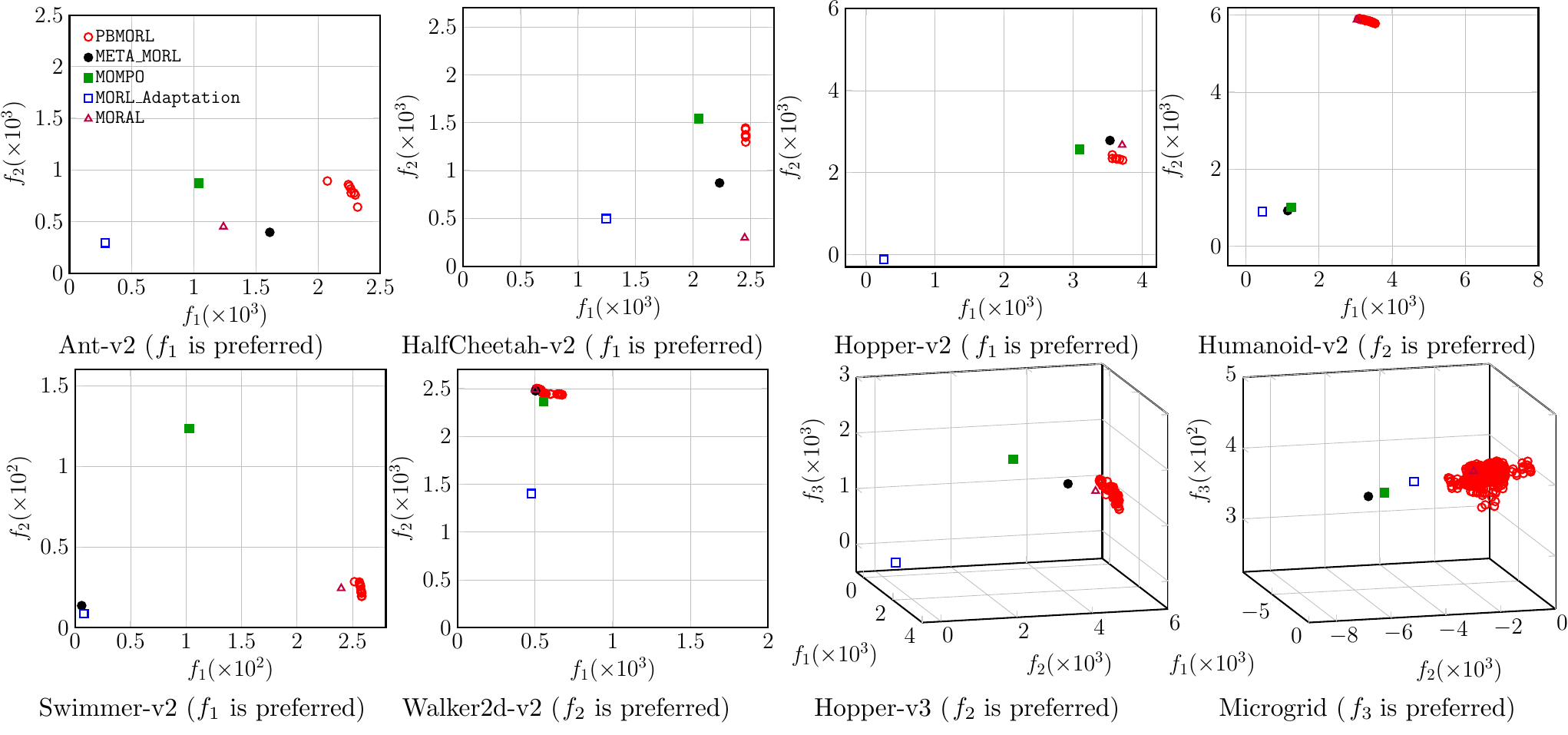}
    \caption{Selected plots of non-dominated policies obtained by \our\ vs \texttt{MORL-Adaptation}, \texttt{META-MORL}, \texttt{MOMPO} and \texttt{MORAL}.}
    \label{fig:preference} 
\end{figure*}

\begin{table*}[t!]
  \centering
  \caption{Comparison results of $\epsilon^\star(\Pi)$ of \our\ versus \texttt{MOMPO}, \texttt{META-MORL}, \texttt{MORL-Adaptation} and \texttt{MORAL} over $10$ runs with mean and standard deviation.}
  \resizebox{.8\linewidth}{!}{ 
    \begin{tabular}{c|c|ccccc}
      \toprule    \multicolumn{1}{c}{\multirow{2}[4]{*}{}}&       & \multicolumn{1}{c}{\texttt{PBMORL}} & \multicolumn{1}{c}{\texttt{MOMPO}} & \multicolumn{1}{c}{\texttt{META-MORL}} & \multicolumn{1}{c}{\texttt{MORL-Adaptation}} & \multicolumn{1}{c}{\texttt{MORAL}} \\
      \midrule
      \multirow{1}[4]{*}{\texttt{Ant-v2}} & $f_1$    & \cellcolor[rgb]{ .702,  .702,  .702}{$\mathbf{7.709(1.68E-3)}$} & $8.910$($1.81$E$-2$) & $8.355$($1.36$E$-2$) & $9.686$($8.53$E$-3$) & $8.926$($1.18$E$-2$) \\
      \cmidrule{2-7}    & $f_2$    & \cellcolor[rgb]{ .702,  .702,  .702}{$\mathbf{7.637(6.98E-4)}$} & $8.914$($2.05$E$-3$) & $8.446$($4.52$E$-3$)  & $9.633$($3.37$E$-3$)  & $8.377$($2.49$E$-3$) \\
      \midrule
    \multirow{1}[4]{*}{\texttt{HalfCheetah-v2}} & $f_1$    & \cellcolor[rgb]{ .702,  .702,  .702}{$\mathbf{7.541(3.21E-6)}$} & $7.953$($1.05$E$-4$)  & $7.752$($1.77$E$-4$)  & $8.743$($1.52$E$-4$) & $7.545$($9.73$E$-5$) \\
    \cmidrule{2-7}    & $f_2$    & \cellcolor[rgb]{ .702,  .702,  .702}{$\mathbf{7.500(3.70E-9)}$} & \cellcolor[rgb]{ .702,  .702,  .702}{$\mathbf{7.500(7.72E-8)}$}  & $7.980$($4.57$E$-8$)  & $8.561$($1.55$E$-7$) & \cellcolor[rgb]{ .702,  .702,  .702}{$\mathbf{7.500(5.66E-8)}$} \\
      \midrule
    \multirow{1}[4]{*}{\texttt{Hopper-v2}} & $f_1$    & \cellcolor[rgb]{ .702,  .702,  .702}{$\mathbf{6.221(1.14E-2)}$} & $6.828$($3.60$E$-3$)  & $6.404$($3.54$E$-3$) & $9.662$($2.32$E$-3$) & $6.236$($2.42$E$-3$)  \\
    \cmidrule{2-7}    & $f_2$    & \cellcolor[rgb]{ .702,  .702,  .702}{$\mathbf{4.587(2.250E-2)}$} & $8.583$($2.31$E$-2$)  & $4.930$($4.68$E$-1$)  & $9.942$($6.12$E$-2$)   & $9.761$($7.94$E$-1$) \\
      \midrule
    \multirow{1}[4]{*}{\texttt{Humanoid-v2}} & $f_1$    & \cellcolor[rgb]{ .702,  .702,  .702}{$\mathbf{2.362(2.51E-2)}$} & $9.292$($4.17$E$-1$)  & $8.679$($5.18$E$-2$) & $9.554$($8.33$E$-2$)  & $5.599$($6.40$E$-2$) \\
    \cmidrule{2-7}    & $f_2$    & \cellcolor[rgb]{ .702,  .702,  .702}{$\mathbf{4.099(6.96E-7)}$} & $8.993$($1.84$E$-2$) & $9.083$($2.84$E$-3$)  & $9.912$($8.34$E$-2$) & $9.776$($1.34$E$-2$) \\
      \midrule
    \multirow{1}[4]{*}{\texttt{Swimmer-v2}} & $f_1$    & \cellcolor[rgb]{ .702,  .702,  .702}{$\mathbf{9.747(1.12E-4)}$} & $9.913$($6.88$E$-4$) & $9.933$($5.58$E$-3$)  & $9.941$($1.87$E$-3$)  & $9.759$($2.99$E$-5$) \\
    \cmidrule{2-7}    & $f_2$    & \cellcolor[rgb]{ .702,  .702,  .702}{$\mathbf{9.850(1.00E-12)}$} & \cellcolor[rgb]{ .702,  .702,  .702}{$\mathbf{9.850(2.87E-9)}$} & $9.860$($3.34$E$-7$) & \cellcolor[rgb]{ .702,  .702,  .702}{$\mathbf{9.850(1.09E-8)}$}  & \cellcolor[rgb]{ .702,  .702,  .702}{$\mathbf{9.850(7.22E-9)}$} \\
      \midrule   
    \multirow{1}[4]{*}{\texttt{Walker2d-v2}} & $f_1$    & \cellcolor[rgb]{ .702,  .702,  .702}{$\mathbf{8.048(9.77E-4)}$} & $8.905$($5.78$E$-3$)  & $8.267$($1.26$E$-2$)   & $9.271$($4.22$E$-3$)  & $8.850($6.31$E$-3$)$ \\
    \cmidrule{2-7}    & $f_2$    & \cellcolor[rgb]{ .702,  .702,  .702}{$\mathbf{7.500(3.21E-8)}$} & $7.643$($6.78$E$-3$)  & $7.528$($2.39$E$-6$)  & $8.602$($6.61$E$-3$)  & \cellcolor[rgb]{ .702,  .702,  .702}{$\mathbf{7.500(1.92E-7)}$} \\
      \midrule
    \multirow{2}[6]{*}{\texttt{Hopper-v3}} & $f_1$    & \cellcolor[rgb]{ .702,  .702,  .702}{$\mathbf{6.020(8.91E-8)}$} & $6.821$($2.91$E$-7$)  & $6.773$($3.51$E$-5$)  & $9.070$($8.77$E$-4$)  & $6.391$($6.98$E$-6$) \\
    \cmidrule{2-7}    & $f_2$    & \cellcolor[rgb]{ .702,  .702,  .702}{$\mathbf{4.346(2.02E-3)}$} & $7.295$($1.72$E$-3$)  & $6.032$($3.55$E$-3$)  & $9.411$($3.87$E$-2$) & $5.263$($9.21$E$-3$) \\
    \cmidrule{2-7}    & $f_3$    & \cellcolor[rgb]{ .702,  .702,  .702}{$\mathbf{7.500(1.90E-8)}$} & \cellcolor[rgb]{ .702,  .702,  .702}{$\mathbf{7.500(2.63E-6)}$}  & $7.522$($3.96$E$-6$) & \cellcolor[rgb]{ .702,  .702,  .702}{$\mathbf{7.500(5.47E-7)}$}  & \cellcolor[rgb]{ .702,  .702,  .702}{$\mathbf{7.500(4.54E-7)}$} \\
      \midrule
    \multirow{2}[6]{*}{\texttt{Microgrid}} & $f_1$    & \cellcolor[rgb]{ .702,  .702,  .702}{$\mathbf{1.149(8.85E-4)}$} & $2.418$($3.39$E$-3$)   & $2.737$($9.05$E$-3$)   & $4.69$($7.22$E$-3$)   & $2.48$($6.72$E$-4$) \\
    \cmidrule{2-7}    & $f_2$    & \cellcolor[rgb]{ .702,  .702,  .702}{$\mathbf{0.000(4.87E-7)}$} & $1.366$($9.02$E$-4$)   & $1.317$($8.62$E$-4$)  & $7.020$($2.31$E$-4$)   & $0.672$($5.56$E$-5$) \\
    \cmidrule{2-7}    & $f_3$    & \cellcolor[rgb]{ .702,  .702,  .702}{$\mathbf{0.030(9.78E-8)}$} & $0.102$($4.57$E$-6$)  & $0.071$($8.76$E$-5$)  & $0.244$($9.94$E$-4$)  & $0.124$($1.29$E$-5$) \\
      \bottomrule
    \end{tabular}
  }
  \label{tab:with_preference}
\end{table*}

Next, we compare \our\ with four other preference-based MORL algorithms. Note that since these peer algorithms are designed to find only one preferred policy, we only evaluate the performance on $\epsilon^\star(\Pi)$ and the corresponding comparison results are shown in~\pref{tab:with_preference}. From these results, we can see that \our\ is the most competitive algorithm in all test cases, though it achieves the same $\epsilon^\star(\Pi)$ as some peer algorithms on selected cases. To have a better visual interpretation, like~\pref{sec:rq1}, we plot the non-dominated policies obtained by different algorithms. Due to the page limit, we only show one type of preference while the full results can be found in~\pref{fig:appendix_exp2} of the \textsc{Appendix}. From these plots, we again divide the performance into two categorizes. For the first category, including \texttt{Ant-v2}, \texttt{Hopper-v3}, \texttt{Humanoid-v2}, and \texttt{Microgrid}, it is evident that \our\ consistently outperforms its peers. Notably, all four preference-based MORL algorithms can only find one policy, which is dominated by nearly all trade-off alternative policies identified by \our. For the other problems, the performance of \our\ is comparable to some of the other peers. This can be explained by the fact that the PFs of such problems are relatively easier to approximate, thus allowing for better exploration towards the region of interest therein.

\subsection{Further Analysis}
\label{sec:further_analysis}

The previous experiments demonstrated the effectiveness of \our\ for finding the policies of interest according to the DM's preferences. In this subsection, we plan to further analyze several core algorithmic components by addressing the following five research questions (RQs).
\begin{enumerate}[RQ1:]
    \item When do we consult the DM?
    \item What is impact of the interaction frequency?
    \item What is the benefit of our preference learning?
    \item What is the impact of the parameters $\kappa_1$ and $\kappa_2$ in the \texttt{preference translation} step?
    \item What if the DM's preference is not determinstic?
    \item What if we use other the aggregation function other than the linear aggregation in~\pref{eq:ws}?
\end{enumerate}

\subsubsection{Investigation on RQ1}
\label{sec:rq1}

\begin{figure*}[t!]
	\center
    \includegraphics[width=\linewidth]{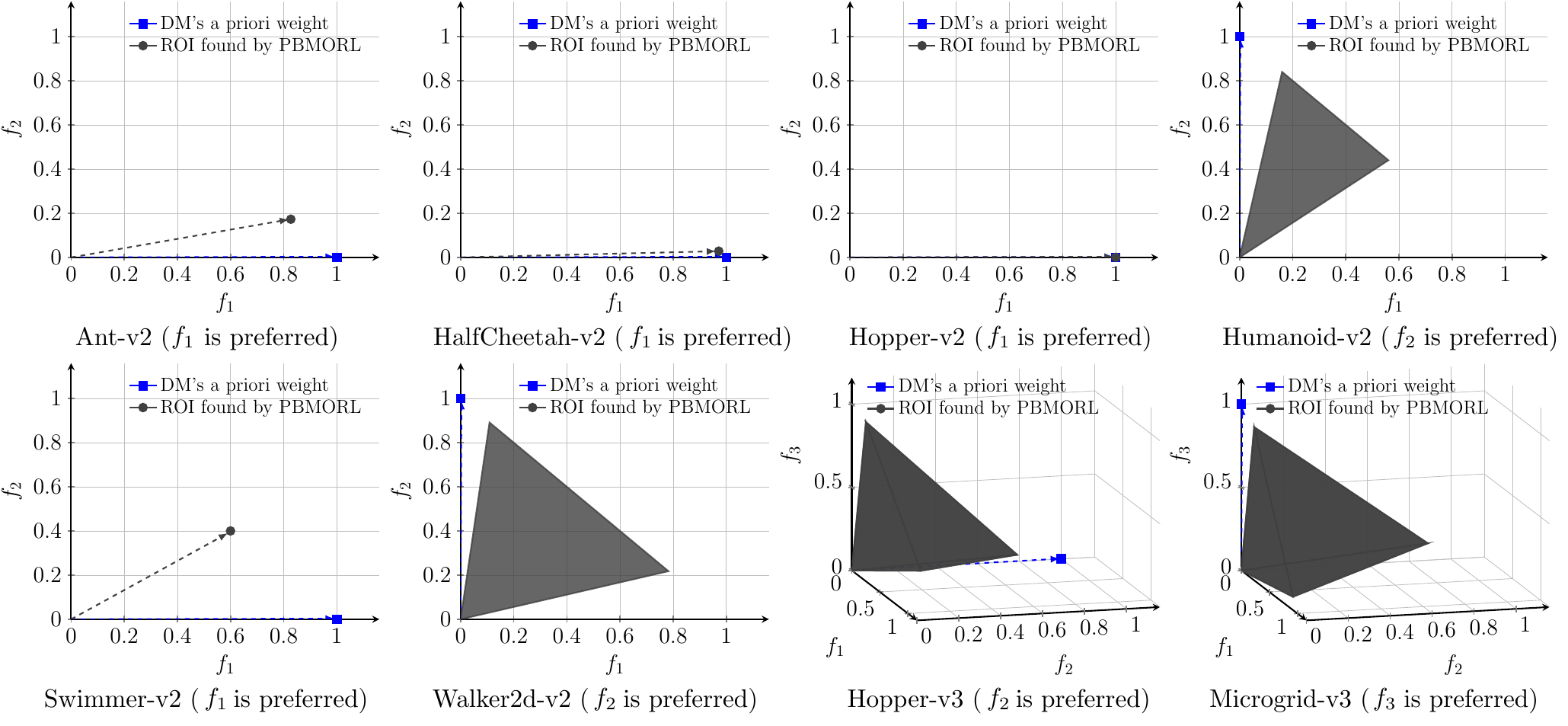}
    \caption{Comparison of the weight vector specified by the DM \textit{a priori} versus the ROI identified by \our\ (shaded in the gray region).}   
    \label{fig:weight_comparison} 
\end{figure*}

In addition to the \textit{interactive} preference elicitation considered in \our, DM's preferences can also be incorporated in \textit{a priori} or \textit{a posteriori} manner. From the comparison results discussed in~\pref{sec:no_preference}, we can see that the a posteriori method may fail to identify the policy of interest in some of the challenging problems. The better performance of \our\ against the other four peer preference-based MORL, as shown in~\pref{sec:with_preference}, demonstrate that the interactive MORL outperforms the a priori method. To have a better understanding of the discrepancy between the a priori and interactive preference elicitation, we plot the corresponding weight vector a priori specified by the DM versus those identified by \our, while the full results can be found in~\pref{fig:appendix_exp31} of the \textsc{Appendix}. From these plots, we can see that the weight vector specified by the DM is always outside the region of interest (ROI). From the DM's perspective, it is not intuitive for the DM to elicit an appropriate weight vector a priori given the black-box nature of the problem itself. Our experiments in~\pref{sec:rq1} demonstrate that there is even no guarantee to find good non-dominated policies without considering the DM's preference information before a posteriori decision-making.

\subsubsection{Investigation on RQ2}
\label{sec:rq2}

\begin{figure*}[t!]
	\center
    \includegraphics[width=\linewidth]{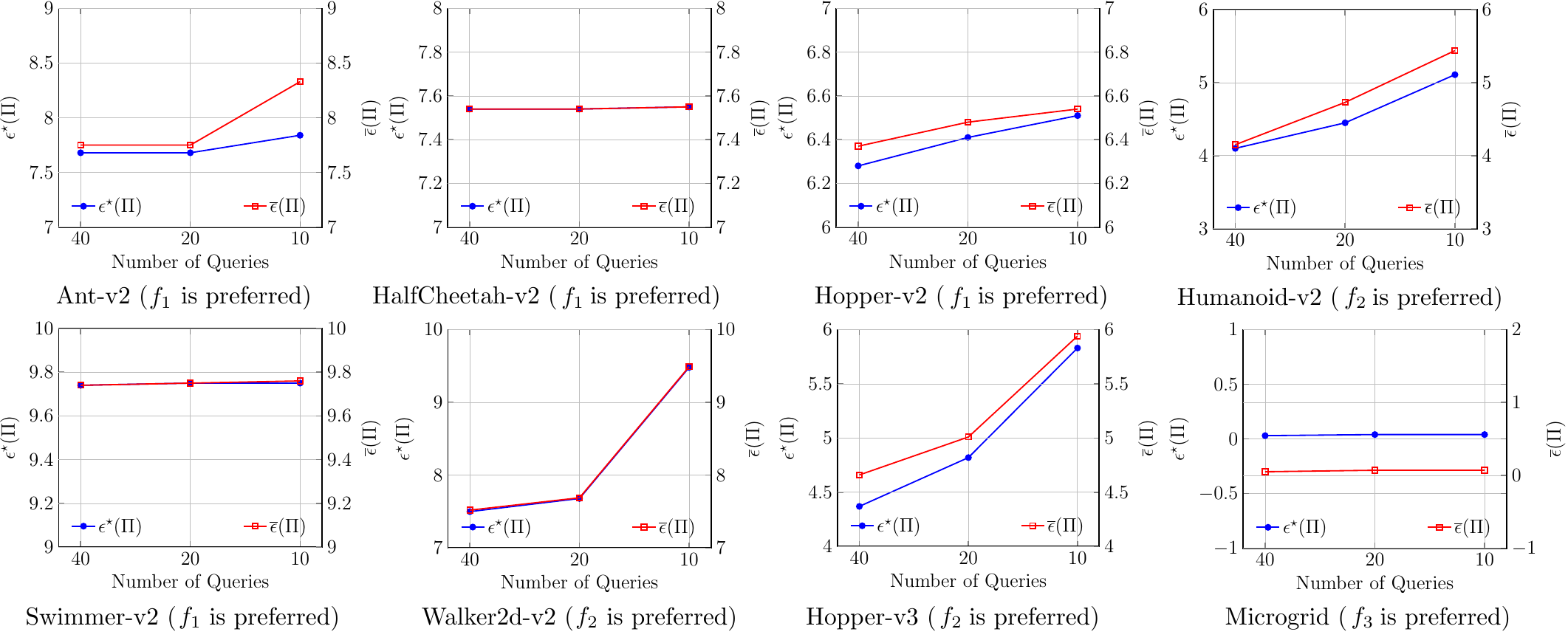}
    \caption{Comparison results of $\epsilon^\star(\Pi)$ and $\bar{\epsilon}(\Pi)$ obtained by \our\ with $10$, $20$, and $40$ interactions, respectively.}
    \label{fig:number_queries} 
\end{figure*}

In the \texttt{consultation} stage, more frequent interactions with the DM would generate more oracles for preference learning. However, this also significantly increases the DM's workload. To address RQ2, we set the number of interactions to $20$ and compared this with $40$ and $10$ consultations. From the selected results shown in~\pref{fig:number_queries}, while full results can be found in~\pref{fig:appendix_exp32} of the \textsc{Appendix}, we observe that reducing the number of interactions does not negatively impact \our's performance on some problems like \texttt{HalfCheetah-v2} with a preference for $f_2$. However, for other problems like \texttt{Walker2d-v2} with a preference for $f_2$, decreasing the number of queries to $10$ can significantly hurt \our's performance. On the other hand, increasing the number of interactions does not bring significantly better performance but can increase the DM's cognitive load.

\subsubsection{Investigation on RQ3}
\label{sec:rq3}

In principle, any off-the-shelf preference learning methods can be used for modeling the DM's preference information. To validate this assertion, we construct two variants of \our, dubbed \texttt{RSMORL} and \texttt{WLMORL}. They replace the preference learning method introduced in~\pref{sec:preference_model} with a classic preference learning approach ranking-SVM~\cite{Joachims02} and a preference learning method in RL~\cite{WirthFN16}, respectively. Note that all variants apply one query at each round of policy optimization, as done in \our. The comparison results shown in~\pref{tab:with_variants} demonstrate the better performance of \our\ on most problems. Specifically, comparing to \texttt{RSMORL}, our better performance not only highlights the superiority of GP for preference learning, but also showcases the benefits of uncertainty provided by GP prediction for better exploration. Comparing to \texttt{WLMORL} as well as \texttt{MORAL}, our better performance demonstrates that learning preferred weight vector(s) is less reliable than learning the DM's preference in the objective space.

\begin{table*}[t!]
    \centering
    \caption{Comparison results of $\epsilon^\star(\Pi)$ and $\overline{\epsilon}(\Pi)$ of \our\ versus \texttt{RSMORL} and \texttt{WLMORL} over $10$ runs with mean and standard deviation.}
    \resizebox{.8\linewidth}{!}{ 
    \begin{tabular}{c|c|c|c|c|c|c|c}
        \toprule    \multicolumn{1}{c}{\multirow{2}[4]{*}{}}&       & \multicolumn{2}{c|}{\texttt{PBMORL}} & \multicolumn{2}{c|}{\texttt{RSMORL}} & \multicolumn{2}{c}{\texttt{WLMORL}} \\
        \cmidrule{3-8}    \multicolumn{1}{c}{} & & $\epsilon^\star(\Pi)$ & $\overline{\epsilon}(\Pi)$ &  $\epsilon^\star(\Pi)$ & $\overline{\epsilon}(\Pi)$ & $\epsilon^\star(\Pi)$ & $\overline{\epsilon}(\Pi)$\\
        \midrule
        \multirow{1}[4]{*}{\texttt{Ant-v2}} & $f_1$    & \cellcolor[rgb]{ .702, .702, .702}{$\mathbf{7.709(1.68E-3)}$} &\cellcolor[rgb]{ .702, .702, .702}{$\mathbf{7.880(3.32E-2)}$} & $8.333$($1.44$E$-5$)  & $8.796$($6.22$E$-3$)  & $7.746$($1.35$E$-5$)  & $7.949$($5.02$E$-3$) \\
        \cmidrule{2-8}    & $f_2$    & \cellcolor[rgb]{ .702, .702, .702}{$\mathbf{7.637(6.98E-4)}$} &\cellcolor[rgb]{ .702, .702, .702}{$\mathbf{7.751(1.65E-4)}$} & $7.722$($1.24$E$-5$)  & $7.998$($5.72$E$-3$)  & $7.695$($8.74$E$-7$)  & $7.956$($2.23$E$-4$) \\
        \midrule
        \multirow{1}[4]{*}{\texttt{HalfCheetah-v2}} & $f_1$    & \cellcolor[rgb]{ .702, .702, .702}{$\mathbf{7.541(3.21E-6)}$} &\cellcolor[rgb]{ .702, .702, .702}{$\mathbf{7.542(7.53E-6)}$} & $7.566$($3.71$E$-4$)  & $7.721$($5.29$E$-4$)  & $8.397$($1.31$E$-4$)  & $9.233$($8.53$E$-2$) \\
        \cmidrule{2-8}    & $f_2$    & \cellcolor[rgb]{ .702, .702, .702}{$\mathbf{7.500(3.70E-9)}$} &\cellcolor[rgb]{ .702, .702, .702}{$\mathbf{7.508(1.17E-4)}$} & \cellcolor[rgb]{ .702, .702, .702}$\mathbf{7.500(2.47E-7)}$ & $7.541$($2.14$E$-6$)  & \cellcolor[rgb]{ .702, .702, .702}$\mathbf{7.508(7.98E-3)}$ & $7.544$($8.98$E$-2$) \\
        \midrule    
        \multirow{1}[4]{*}{\texttt{Hopper-v2}} & $f_1$    & \cellcolor[rgb]{ .702, .702, .702}{$\mathbf{6.221(1.14E-2)}$} &\cellcolor[rgb]{ .702, .702, .702}{$\mathbf{6.299(1.57E-2)}$} &  $9.566$($8.17$E$-5$)  & $9.642$($5.58$E$-2$) & $9.421$($6.43$E$-7$)  & $9.631$($5.16$E$-3$) \\
        \cmidrule{2-8}    & $f_2$    & \cellcolor[rgb]{ .702, .702, .702}{$\mathbf{4.587(2.25E-2)}$} &\cellcolor[rgb]{ .702, .702, .702}{$\mathbf{4.972(3.11E-2)}$} & $8.351$($5.29$E$-5$)  & $8.635$($3.68$E$-2$)  & $9.922$($3.11$E$-5$)  & $9.730$($6.21$E$-3$) \\
        \midrule
        \multirow{1}[4]{*}{\texttt{Humanoid-v2}} & $f_1$    & \cellcolor[rgb]{ .702, .702, .702}{$\mathbf{2.362(2.51E-2)}$} &\cellcolor[rgb]{ .702, .702, .702}{$\mathbf{2.583(8.31E-1)}$} & $6.324$($5.21$E$-5$)  & $6.543$($5.19$E$-3$)  & $9.236$($5.66$E$-5$)  & $9.353$($6.74$E$-2$) \\
        \cmidrule{2-8}    & $f_2$    & \cellcolor[rgb]{ .702, .702, .702}{$\mathbf{4.099(6.96E-7)}$} &\cellcolor[rgb]{ .702, .702, .702}{$\mathbf{4.143(9.79E-5)}$} & $4.818$($5.89$E$-5$)  & $5.076$($2.29$E$-2$)  & $8.718$($1.39$E$-6$)  & $8.728$($5.21$E$-2$) \\
        \midrule    
        \multirow{1}[4]{*}{\texttt{Swimmer-v2}} & $f_1$    & \cellcolor[rgb]{ .702, .702, .702}{$\mathbf{9.747(1.12E-4)}$} & \cellcolor[rgb]{ .702, .702, .702}{$\mathbf{9.749(1.49E-4)}$} & $9.716$($6.23$E$-5$)  & $9.762$($7.22$E$-3$)  & $9.884$($7.22$E$-3$)  & $9.894$($2.39$E$-3$) \\
        \cmidrule{2-8}    & $f_2$    & \cellcolor[rgb]{ .702, .702, .702}{$\mathbf{9.850(1.00E-12)}$}  &\cellcolor[rgb]{ .702, .702, .702}{$\mathbf{9.850(3.67E-12)}$} & \cellcolor[rgb]{ .702, .702, .702}$\mathbf{9.850(5.39E-8)}$ & \cellcolor[rgb]{ .702, .702, .702}$\mathbf{9.850(1.12E-7)}$ & \cellcolor[rgb]{ .702, .702, .702}$\mathbf{9.850(8.84E-7)}$ & \cellcolor[rgb]{ .702, .702, .702}$\mathbf{9.850(6.23E-6)}$ \\
        \midrule    
        \multirow{1}[4]{*}{\texttt{Walker2d-v2}} & $f_1$    & \cellcolor[rgb]{ .702, .702, .702}{$\mathbf{8.048(9.77E-4)}$} & \cellcolor[rgb]{ .702, .702, .702}{$\mathbf{8.116(3.95E-3)}$} &  $8.636$($5.13$E$-5$)  & $8.928$($7.11$E$-2$)  & $8.609$($4.26$E$-5$) & $8.897$($9.21$E$-3$) \\
        \cmidrule{2-8}    & $f_2$    & \cellcolor[rgb]{ .702, .702, .702}{$\mathbf{7.500(3.21E-8)}$}  &\cellcolor[rgb]{ .702, .702, .702}{$\mathbf{7.506(8.84E-5)}$} & \cellcolor[rgb]{ .702, .702, .702}$\mathbf{7.500(6.64E-6)}$ & $7.534$($8.34$E$-4$)  & $7.600$($1.14$E$-5$)   & $7.673$($9.96$E$-3$) \\
        \midrule    
        \multirow{1}[6]{*}{\texttt{Hopper-v3}} & $f_1$    & $6.020$($8.91$E$-8$)  &\cellcolor[rgb]{ .702, .702, .702}{$\mathbf{6.203(1.44E-3)}$} & \cellcolor[rgb]{ .702, .702, .702}$\mathbf{5.981(5.12E-5)}$  & $6.250$($9.75$E$-3$) & $6.858$($6.75$E$-6$)  & $7.736$($7.34$E$-3$) \\
        \cmidrule{2-8}    & $f_2$    & \cellcolor[rgb]{ .702, .702, .702}{$\mathbf{4.346(2.02E-3)}$} & \cellcolor[rgb]{ .702, .702, .702}{$\mathbf{4.539(1.97E-2)}$} & $5.000$($2.66$E$-5$)     & $5.272$($9.74$E$-4$)  & $8.611$($1.74$E$-6$)  & $9.657$($8.72$E$-4$) \\
        \cmidrule{2-8}    & $f_3$    & \cellcolor[rgb]{ .702, .702, .702}{$\mathbf{7.500(1.90E-8)}$}  &\cellcolor[rgb]{ .702, .702, .702}{$\mathbf{7.502(1.23E-5)}$} & \cellcolor[rgb]{ .702, .702, .702}$\mathbf{7.500(2.74E-4)}$ & $7.510$($8.82$E$-3$)  & \cellcolor[rgb]{ .702, .702, .702}$\mathbf{7.500(1.08E-6)}$ & \cellcolor[rgb]{ .702, .702, .702}$\mathbf{7.500(4.34E-3)}$ \\
        \midrule    
        \multirow{1}[6]{*}{\texttt{Microgrid}} & $f_1$    & \cellcolor[rgb]{ .702, .702, .702}{$\mathbf{1.149(8.85E-4)}$} & \cellcolor[rgb]{ .702, .702, .702}{$\mathbf{1.846(3.88E-5)}$} & \cellcolor[rgb]{ .702, .702, .702}$\mathbf{1.149(1.34E-3)}$ & $2.382$($8.87$E$-2$)  & \cellcolor[rgb]{ .702, .702, .702}$\mathbf{1.149(5.88E-5)}$ & $2.397$($1.74$E$-3$) \\
        \cmidrule{2-8}    & $f_2$    & \cellcolor[rgb]{ .702, .702, .702}{$\mathbf{0.000(4.87E-7)}$} &\cellcolor[rgb]{ .702, .702, .702}{$\mathbf{0.511(3.41E-6)}$} & \cellcolor[rgb]{ .702, .702, .702}$\mathbf{0.000(2.62E-5)}$ & $1.558$($6.62$E$-3$)  & \cellcolor[rgb]{ .702, .702, .702}$\mathbf{0.000(2.34E-5)}$ & $1.045$($8.78$E$-4$) \\
        \cmidrule{2-8}    & $f_3$    &\cellcolor[rgb]{ .702, .702, .702}{$\mathbf{0.030(9.78E-8)}$}  &\cellcolor[rgb]{ .702, .702, .702}{$\mathbf{0.050(5.74E-8)}$} & \cellcolor[rgb]{ .702, .702, .702}$\mathbf{0.030(1.22E-5)}$ & $0.081$($8.87$E$-7$)  & \cellcolor[rgb]{ .702, .702, .702}$\mathbf{0.030(3.54E-6)}$ & $0.074$($3.79$E$-4$) \\
        \bottomrule
    \end{tabular}
    }
    \label{tab:with_variants}
\end{table*}

\subsubsection{Investigation on RQ4}
\label{sec:rq4}

In the \texttt{preference translation} step, there are two hyperparameters $\kappa_1$ and $\kappa_2$ that implicitly control the balance between exploration versus exploitation. Specifically, a larger $\kappa_1$ indicates a greater reliance on learned preference information to guide policy optimization, while a larger $\kappa_2$ emphasizes random exploration. We set $\kappa_1=80\%\times|\Pi|$ and $\kappa_2=20\%\times|\Pi|$ as the default. To address RQ4, here we empirically compare the the default setting with three other $\kappa_1\in\{100\%\times|\Pi|, 50\%\times|\Pi|,20\%\times|\Pi|\}$ on three example problems, including \texttt{Walker2d-v2}, \texttt{HalfCheetah-v2}, and \texttt{Swimmer-v2}. From the comparison results in~\pref{tab:kappa}, we find that the setting with $\kappa_1=80\%\times|\Pi|$ and $\kappa_2=20\%\times|\Pi|$ consistently achieves the best performance when considering $\epsilon^\star(\Pi)$. On the other hand, when considering $\overline{\epsilon}(\Pi)$, the outcomes depend more on the characteristics of the individual problems. In general, we find that a large $\kappa_1$ and a nonzero $\kappa_2$ are efficient for most situations.

\begin{table*}[t!]
    \centering
    \caption{Comparison results of $\epsilon^\star(\Pi)$ and $\overline{\epsilon}(\Pi)$ on different settings of $\kappa_1$ and $\kappa_2$ over $10$ runs with mean and standard deviation.}
    \resizebox{\linewidth}{!}{
        \begin{tabular}{c|c|c|c|c|c|c|c|c|c}
            \toprule    \multicolumn{1}{c}{\multirow{2}[4]{*}{}}&       & \multicolumn{2}{c|}{$\kappa_1=100\%\times|\Pi|,\kappa_2=0$} & \multicolumn{2}{c|}{$\kappa_1=80\%\times|\Pi|,\kappa_2=20\%\times|\Pi|$} & \multicolumn{2}{c|}{$\kappa_1=50\%\times|\Pi|,\kappa_2=50\%\times|\Pi|$} & \multicolumn{2}{c}{$\kappa_1=20\%\times|\Pi|,\kappa_2=80\%\times|\Pi|$}\\
            \cmidrule{3-10}    \multicolumn{1}{c}{} & & $\epsilon^\star(\Pi)$ & $\overline{\epsilon}(\Pi)$ &  $\epsilon^\star(\Pi)$ & $\overline{\epsilon}(\Pi)$ & $\epsilon^\star(\Pi)$ & $\overline{\epsilon}(\Pi)$& $\epsilon^\star(\Pi)$ & $\overline{\epsilon}(\Pi)$\\
            \midrule
            \multirow{1}[4]{*}{\texttt{HalfCheetah-v2}} & $f_1$    & \cellcolor[rgb]{ .702, .702, .702}$\mathbf{7.541(7.65E-8)}$ & \cellcolor[rgb]{ .702, .702, .702}$\mathbf{7.542(6.34E-7)}$  & \cellcolor[rgb]{ .702, .702, .702}{$\mathbf{7.541(3.21E-6)}$} &\cellcolor[rgb]{ .702, .702, .702}{$\mathbf{7.542(7.53E-6)}$}  & $7.550($6.24$E$-8$)$  & $7.554$($7.78$E$-7$)&$7.551$($8.75$E$-8$)&$7.562$($2.56$E$-7$) \\
            \cmidrule{2-8}    & $f_2$    & \cellcolor[rgb]{ .702, .702, .702}$\mathbf{7.500(6.76E-10)}$ & \cellcolor[rgb]{ .702, .702, .702}$\mathbf{7.508(3.19E-9)}$  & \cellcolor[rgb]{ .702, .702, .702}{$\mathbf{7.500(3.70E-9)}$} &\cellcolor[rgb]{ .702, .702, .702}{$\mathbf{7.508(1.17E-8)}$}  & \cellcolor[rgb]{ .702, .702, .702}$\mathbf{7.500(4.42E-9)}$  & $7.522$($8.12$E$-4$)&\cellcolor[rgb]{ .702, .702, .702}$\mathbf{7.500(6.24E-6)}$ & $7.534$($3.28$E$-6$) \\
            \midrule
            \multirow{1}[4]{*}{\texttt{Swimmer-v2}} & $f_1$    & $9.761$($6.79$E$-5$) & $9.762$($3.78$E$-4$)  & \cellcolor[rgb]{ .702, .702, .702}{$\mathbf{9.747(1.12E-4)}$} & \cellcolor[rgb]{ .702, .702, .702}{$\mathbf{9.749(1.49E-4)}$}  & $9.770$($5.63$E$-4$)  & $9.773$($3.29$E$-5$) & $9.763$($6.79$E$-5$)  & $9.768$($1.12$E$-5$) \\
            \cmidrule{2-8}    & $f_2$    & \cellcolor[rgb]{ .702, .702, .702}$\mathbf{9.850(4.55E-10)}$ & \cellcolor[rgb]{ .702, .702, .702}$\mathbf{9.850(7.87E-11)}$  & \cellcolor[rgb]{ .702, .702, .702}{$\mathbf{9.850(1.00E-12)}$}  &\cellcolor[rgb]{ .702, .702, .702}{$\mathbf{9.850(3.67E-12)}$}  & \cellcolor[rgb]{ .702, .702, .702}$\mathbf{9.850(2.56E-10)}$ & \cellcolor[rgb]{ .702, .702, .702}$\mathbf{9.850(7.73E-10)}$ & \cellcolor[rgb]{ .702, .702, .702}$\mathbf{9.850(6.66E-9)}$ & \cellcolor[rgb]{ .702, .702, .702}$\mathbf{9.850(8.34E-9)}$ \\
            \midrule  
            \multirow{1}[4]{*}{\texttt{Walker2d-v2}} & $f_1$    & $8.612$($2.37$E$-4$) & $8.652$($6.17$E$-4$)  & \cellcolor[rgb]{ .702, .702, .702}{$\mathbf{8.048(9.77E-4)}$} & \cellcolor[rgb]{ .702, .702, .702}{$\mathbf{8.116(3.95E-3)}$}  & $8.623$($8.13$E$-3$)  & $8.738$($6.32$E$-2$) & $8.513$($7.93$E$-4$) & $8.671$($5.35$E$-3$) \\
            \cmidrule{2-8}    & $f_2$    & \cellcolor[rgb]{ .702, .702, .702}$\mathbf{7.500(6.61E-6)}$ & $7.593$($6.25$E$-5$)  & \cellcolor[rgb]{ .702, .702, .702}{$\mathbf{7.500(3.21E-8)}$}  &\cellcolor[rgb]{ .702, .702, .702}{$\mathbf{7.506(8.84E-5)}$}  & \cellcolor[rgb]{ .702, .702, .702}$\mathbf{7.500(7.27E-6)}$  & $7.510$($3.90$E$-4$)& \cellcolor[rgb]{ .702, .702, .702}$\mathbf{7.500(1.22E-5)}$  & $7.516$($8.28$E$-4$) \\
            \bottomrule
        \end{tabular}
        }
        \label{tab:kappa}
\end{table*}

\subsubsection{Investigation on RQ5}
\label{sec:rq5}

\begin{figure*}[h]
	\center
    \includegraphics[width=\linewidth]{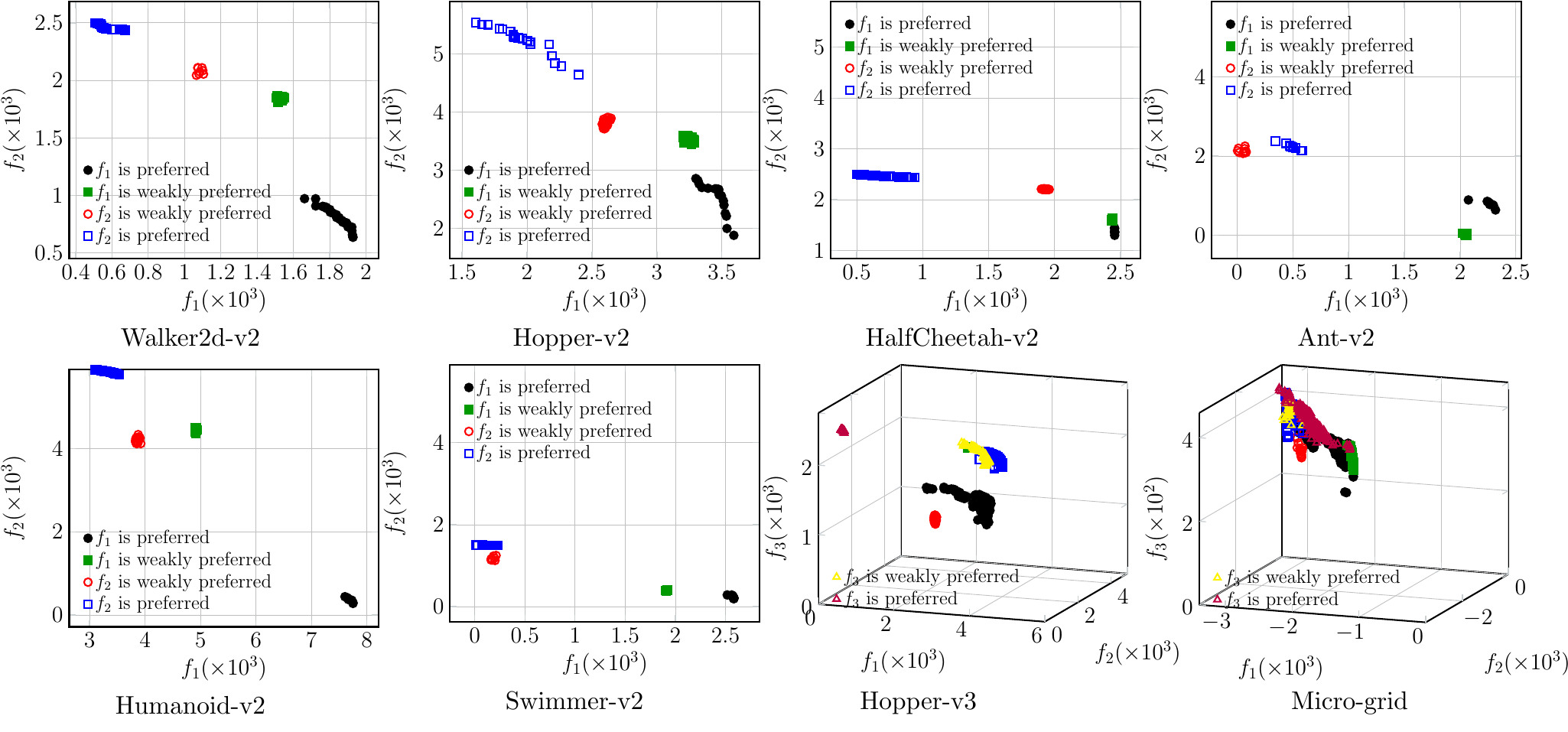}
    \caption{Selected plots of the non-dominated policies obtained by \our\ with different types of preference settings. One is \textit{deterministic} preference on one objective while the other is a \textit{fuzzy} type of preference.}
    \label{fig:fuzzy} 
\end{figure*}

The DM's preferences discussed so far in our experiments are all \textit{deterministic}. That is to say DMs are assumed to prefer only one objective function over the other(s). However, it is not uncommon that DMs can be blurry about their preference. For instance, the DM can be more into one objective (say $70\%$), but she also gives certain level of priority (say the remaining $30\%$) to the other objective(s). Here, we conduct an experiment that considers a \textit{fuzzy} type of preference. From the plots of the non-dominated policies found by different types of preference settings in~\pref{fig:fuzzy}, we find that our proposed \our\ can also be used to find trade-off policies with a controllable bias towards one of the objectives, instead of a polarized preference. However, we also find that the trade-off policies found by \our\ in \texttt{Ant-v2} are further polarized in the less preferred objective. This can be attributed to the intriguing interaction of two objectives in \texttt{Ant-v2} or the difficulty of \our\ when tackling this environment. All in all, it is an interesting and important future direction to investigate more diversified types of preference elicitation methods under the \our\ framework.

\subsubsection{Investigation on RQ6}
\label{sec:rq6}

\begin{figure*}[t!]
	\center
    \includegraphics[width=\linewidth]{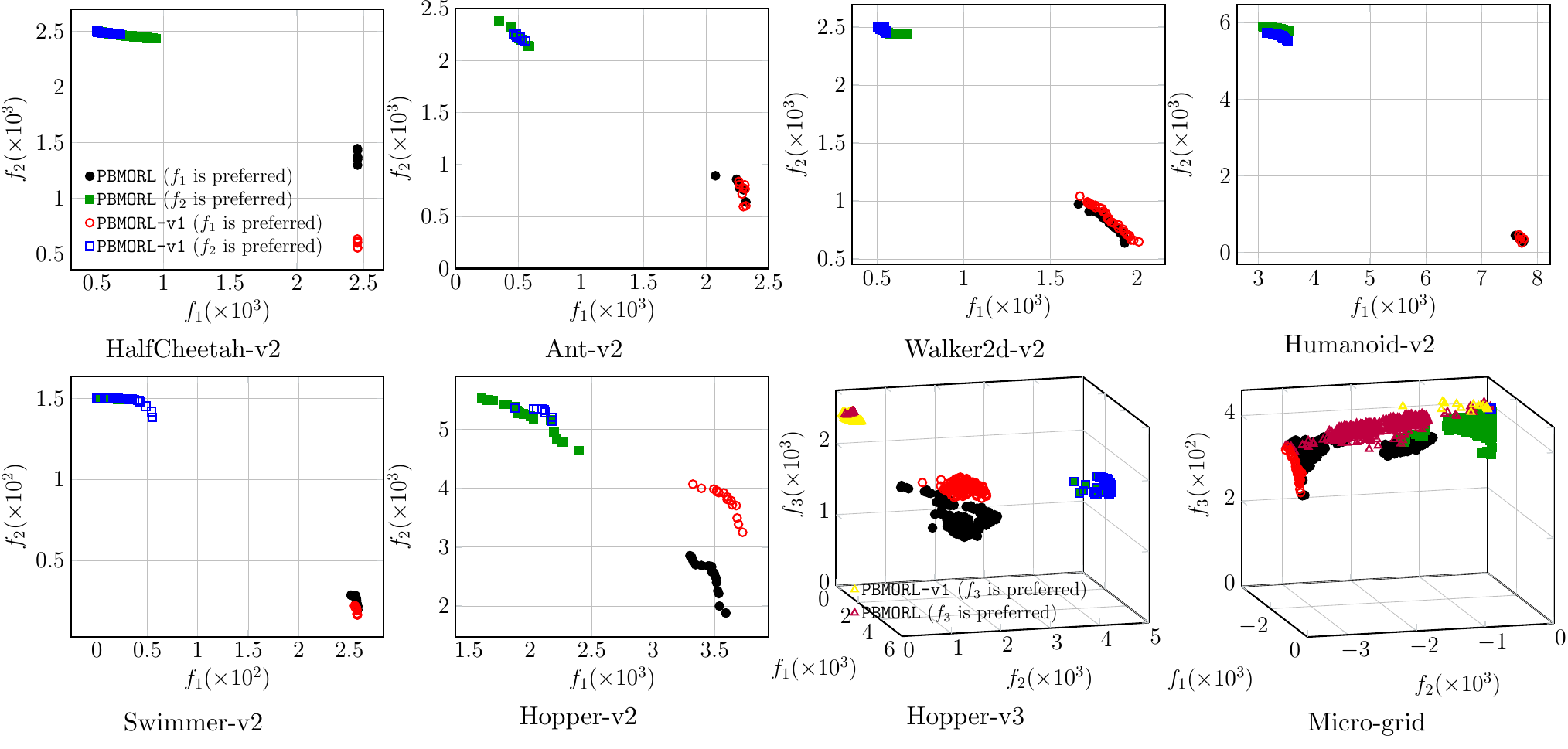}
    \caption{Plots of the non-dominated policies obtained by \our\ versus \texttt{PBMORL-TCH} with different preferences on each objective.}
    \label{fig:tch} 
\end{figure*}

As one of the key components of \our, the \texttt{seeding} module works as a conventional MORL to search for a set of promising trade-off policies that approximate the PF. While we applied a linear aggregation in \our\ for a proof-of-concept purpose, it has been notorious in the multi-objective optimization domain (e.g.,~\cite{deb01,ZhangL07,LiZW21}) that~\pref{eq:ws} can be ineffective to search for solutions located in the non-convex region(s) of the PF. In contrast, the following weighted Tchebycheff aggregation function has been widely recognized to be applicable to problems with both convex and non-convex regions in the PF:
\begin{equation}
    \tilde{J}(\pi,\mathbf{w},\mathbf{z})=\max_{i\in\{1,\ldots,m\}}\left\{w_i|J_{i}(\pi)-z_i|\right\},
    \label{eq:tch}
\end{equation}
where $\mathbf{z}=(z_1,\cdots,z_m)^\top$ is the utopia point. Note that all objective functions in this paper are considered being maximized, whereas there is no \textit{a priori} knowledge about the maximum of each objective function. In this case, we set $\mathbf{z}$ as the nadir point instead, i.e., $z_i=0$, $i\in\{1,\ldots,m\}$. To investigate RQ6, we replace~\pref{eq:ws} as~\pref{eq:tch} in the \texttt{seeding} module of \our\ to constitute a variant dubbed \texttt{PBMORL-TCH}. Note that our proposed \our\ is a general framework that each component can be adapted to any other techniques in a plug-in manner. From the comparison results of $\epsilon^\star(\Pi)$ and $\overline{\epsilon}(\Pi)$ shown in~\pref{tab:tch} along with the non-dominated policies shown in~\pref{fig:tch}, we can see that the performance of \our\ and \texttt{PBMORL-TCH} is close to each other. This can be explained as the PF of the MORL problems considered here are all with convex PFs. The robust performance \texttt{PBMORL-TCH} also provides us confidence to extend our proposed \our\ framework for handling problems in more complex environments.

\begin{table*}[t!]
    \centering
    \caption{Comparison results of $\epsilon^\star(\Pi)$ and $\overline{\epsilon}(\Pi)$ of \our\ against \texttt{PBMORL-TCH} over $10$ runs with mean and standard deviation.}
    \resizebox{.65\linewidth}{!}{ 
    \begin{tabular}{c|c|c|c|c|c}
        \toprule    \multicolumn{1}{c}{\multirow{2}[4]{*}{}}&       & \multicolumn{2}{c|}{\texttt{PBMORL-TCH}} & \multicolumn{2}{c}{\texttt{PBMORL}} \\
        \cmidrule{3-6}    \multicolumn{1}{c}{} & & $\epsilon^\star(\Pi)$ & $\overline{\epsilon}(\Pi)$ &  $\epsilon^\star(\Pi)$ & $\overline{\epsilon}(\Pi)$ \\ 
        \midrule
        \multirow{1}[4]{*}{\texttt{Ant-v2}} & $f_1$&\cellcolor[rgb]{ .702, .702, .702}{$\mathbf{7.738(6.41E-3)}$}  &\cellcolor[rgb]{ .702, .702, .702}{$\mathbf{7.866(5.09E-2)}$} & $7.709$($1.68$E$-3$) &$7.880$($3.32$E$-2$)  \\
        \cmidrule{2-6}    & $f_2$    &$7.700$($3.70$E$-3$)&$7.87$($2.06$E$-2$)   & \cellcolor[rgb]{ .702, .702, .702}{$\mathbf{7.637(6.98E-4)}$} &\cellcolor[rgb]{ .702, .702, .702}{$\mathbf{7.751(1.65E-4)}$}   \\
        \midrule
        \multirow{1}[4]{*}{\texttt{HalfCheetah-v2}} & $f_1$&$7.543$($2.00$E$-6$) &$7.591$($1.06$E$-2$)    &\cellcolor[rgb]{ .702, .702, .702}{$\mathbf{7.541(3.21E-6)}$} &\cellcolor[rgb]{ .702, .702, .702}{$\mathbf{7.542(7.53E-6)}$}   \\
        \cmidrule{2-6}    & $f_2$    &$7.501$($3.23$E$-7$) &$7.510$($2.63$E$-5$)    &\cellcolor[rgb]{ .702, .702, .702}{$\mathbf{7.500(3.70E-9)}$} &\cellcolor[rgb]{ .702, .702, .702}{$\mathbf{7.508(1.17E-4)}$}   \\
        \midrule
        \multirow{1}[4]{*}{\texttt{Hopper-v2}} & $f_1$&\cellcolor[rgb]{ .702, .702, .702}{$\mathbf{6.203(3.71E-3)}$} & $6.301$($1.13$E$-2$)&6.221($1.14$E$-2$)&\cellcolor[rgb]{ .702, .702, .702}{$\mathbf{6.299(1.57E-2)}$} \\
        \cmidrule{2-6}    & $f_2$    &$4.700$($4.34$E$-2$)&\cellcolor[rgb]{ .702, .702, .702}{$\mathbf{4.904(1.04E-1)}$}    &\cellcolor[rgb]{ .702, .702, .702}{$\mathbf{4.587(2.25E-2)}$} & $4.972$($3.11$E$-2$) \\
        \midrule
        \multirow{1}[4]{*}{\texttt{Humanoid-v2}} & $f_1$ &\cellcolor[rgb]{ .702, .702, .702}{$\mathbf{2.230(3.64E-4)}$} & \cellcolor[rgb]{ .702, .702, .702}{$\mathbf{2.289(3.39E-5)}$} & $2.362$($2.51$E$-2$) &$2.583$($1.49$E$-1$)     \\
        \cmidrule{2-6}    & $f_2$ &$4.224$($3.55$E$-3$)&$4.284$($2.72$E$-3$) &\cellcolor[rgb]{ .702, .702, .702}{$\mathbf{4.099(6.96E-7)}$} & \cellcolor[rgb]{ .702, .702, .702}{$\mathbf{4.143(9.79E-5)}$}      \\
        \midrule
        \multirow{1}[4]{*}{\texttt{Swimmer-v2}} & $f_1$&$9.751$($7.69$E$-5$)&$9.752$($1.04$E$-4$)  & \cellcolor[rgb]{ .702, .702, .702}{$\mathbf{9.747(1.12E-4)}$} & \cellcolor[rgb]{ .702, .702, .702}{$\mathbf{9.749(1.49E-4)}$}     \\
        \cmidrule{2-6}    & $f_2$    &\cellcolor[rgb]{ .702, .702, .702}{$\mathbf{9.850(3.13E-12)}$}&\cellcolor[rgb]{ .702, .702, .702}{$\mathbf{9.850(1.04E-7)}$}  &\cellcolor[rgb]{ .702, .702, .702}{$\mathbf{9.850(1.00E-12)}$}  &\cellcolor[rgb]{ .702, .702, .702}{$\mathbf{9.850(3.67E-12)}$} \\
        \midrule
        \multirow{1}[4]{*}{\texttt{Walker2d-v2}} & $f_1$&\cellcolor[rgb]{ .702, .702, .702}{$\mathbf{7.922(1.76E-1)}$} &\cellcolor[rgb]{ .702, .702, .702}{$\mathbf{8.067(1.86E-1)}$} & $8.048$($9.77$E$-4$)&$8.116$($3.95$E$-3$)    \\
        \cmidrule{2-6}    & $f_2$ &\cellcolor[rgb]{ .702, .702, .702}{$\mathbf{7.500(4.74E-6)}$} & $7.525$($3.65$E$-4$) &\cellcolor[rgb]{ .702, .702, .702}{$\mathbf{7.500(3.21E-8)}$}  &\cellcolor[rgb]{ .702, .702, .702}{$\mathbf{7.506(8.84E-5)}$}      \\
        \midrule
        \multirow{2}[6]{*}{\texttt{Hopper-v3}} & $f_1$&$6.360$($5.29$E$-2$)&$6.582$($6.21$E$-2$)   & \cellcolor[rgb]{ .702, .702, .702}{$\mathbf{6.020(8.91E-8)}$}  &\cellcolor[rgb]{ .702, .702, .702}{$\mathbf{6.203(1.44E-3)}$}  \\
        \cmidrule{2-6}    & $f_2$ &\cellcolor[rgb]{ .702, .702, .702}{$\mathbf{4.333(1.58E-1)}$} & \cellcolor[rgb]{ .702, .702, .702}{$\mathbf{4.434(4.21E-4)}$}   &$4.346$($2.02$E$-3$)& $4.539$($1.97$E$-2$)      \\
        \cmidrule{2-6}    & $f_3$ &\cellcolor[rgb]{ .702, .702, .702}{$\mathbf{7.500(5.60E-6)}$} & $7.506$($2.05$E$-5$)   &\cellcolor[rgb]{ .702, .702, .702}{$\mathbf{7.500(1.90E-8)}$}  &\cellcolor[rgb]{ .702, .702, .702}{$\mathbf{7.502(1.23E-5)}$}     \\
        \midrule
        \multirow{2}[6]{*}{\texttt{Microgrid}} & $f_1$    &$1.169$($3.96$E$-1$)&\cellcolor[rgb]{ .702, .702, .702}{$\mathbf{1.206(5.20E-4)}$}&\cellcolor[rgb]{ .702, .702, .702}{$\mathbf{1.149(8.85E-4)}$} &$1.846$($3.88$E$-5$)      \\
        \cmidrule{2-6}    & $f_2$&$0.016$($4.94$E$-4$)&\cellcolor[rgb]{ .702, .702, .702}{$\mathbf{0.086(1.13E-2)}$}&\cellcolor[rgb]{ .702, .702, .702}{$\mathbf{0.000(4.87E-7)}$}&$0.511$($3.41$E$-6$) \\
        \cmidrule{2-6}    & $f_3$&$0.039$($5.00$E$-7$)&\cellcolor[rgb]{ .702, .702, .702}{$\mathbf{0.047(2.45E-5)}$}&\cellcolor[rgb]{ .702, .702, .702}{$\mathbf{0.030(9.78E-8)}$}  &$0.050$($5.74$E$-8$)       \\
        \bottomrule
    \end{tabular}
    }
    \label{tab:tch}
\end{table*}

%% file: conclusion.tex

\vspace{-0.5em}
\section{Conclusion and Future Directions}
\label{sec:conclusion}
\vspace{-0.5em}

This paper proposed a human-in-the-loop framework for preference-based MORL that searches for policies of interest preferred by the DM. This framework proactively learns the DM's preferences in an interactive manner, using the learned preference information to guide policy optimization in MORL. It is worth noting that our proposed \our\ is highly versatile, as all its algorithmic components can be replaced by other related techniques in a plug-in manner. Extensive experiments on the \textsc{MuJoCo} and \textsc{MMSD} environments fully demonstrate the effectiveness of our proposed \our\ for finding the policies of interest.

Human-in-the-loop interactive MORL presents a promising paradigm for realizing human-AI collaboration. However, the field is far from mature, and numerous issues warrant exploration in the future. For instance, this paper assumes that the information provided by MORL is fully understandable by the DM, which may not be realistic, particularly when dealing with more than three objective functions. Developing a human-computer interaction platform and mechanism is essential for enhancing the effectiveness of interactive MORL. Furthermore, we assume that the DM's preferences remain consistent throughout the MORL process. Proactively detecting and adapting to changes in the DM's preferences in dynamic and uncertain environments pose a significant challenge~\cite{ChenLY18,FanLT20,ChenL21a,LiCY23}. Another missing, yet important issue, is how to handle constraints in the context of multiple objectives~\cite{LiCFY19,ShanL21}. Things become even more challenging when the constraints are (partially) unobservable~\cite{LiLL22,WangL24}. It is also worth noting that evolutionary computation and multi-objective optimization have been successfully applied to solve real-world problems, e.g., natural language processing~\cite{YangL23}, neural architecture search~\cite{ChenL23,LyuYWHL23,LyuHYCYLWH23,LyuLHWYL23}, robustness of neural networks~\cite{ZhouLM22a,ZhouLM22b,WilliamsLM23a,WilliamsLM23b,WilliamsLM22,WilliamsL23c}, software engineering~\cite{LiXT19,LiXCWT20,LiuLC20,LiXCT20,LiYV23}, smart grid management~\cite{XuLAZ21,XuLA21,XuLA22}, communication networks~\cite{BillingsleyLMMG19,BillingsleyLMMG20,BillingsleyMLMG20,BillingsleyLMMG21}, machine learning~\cite{CaoKWL12,LiWKC13,LiK14,CaoKWLLK15,WangYLK21}, and visualization~\cite{GaoNL19}. Finally, the explainability of the policies of interest and their implications has rarely been discussed in the literature, representing another area for future investigation.

%% file: appendix.tex

\clearpage

\section*{\textsc{Appendix}}

\begin{figure*}[h]
	\center
    \includegraphics[width=\linewidth]{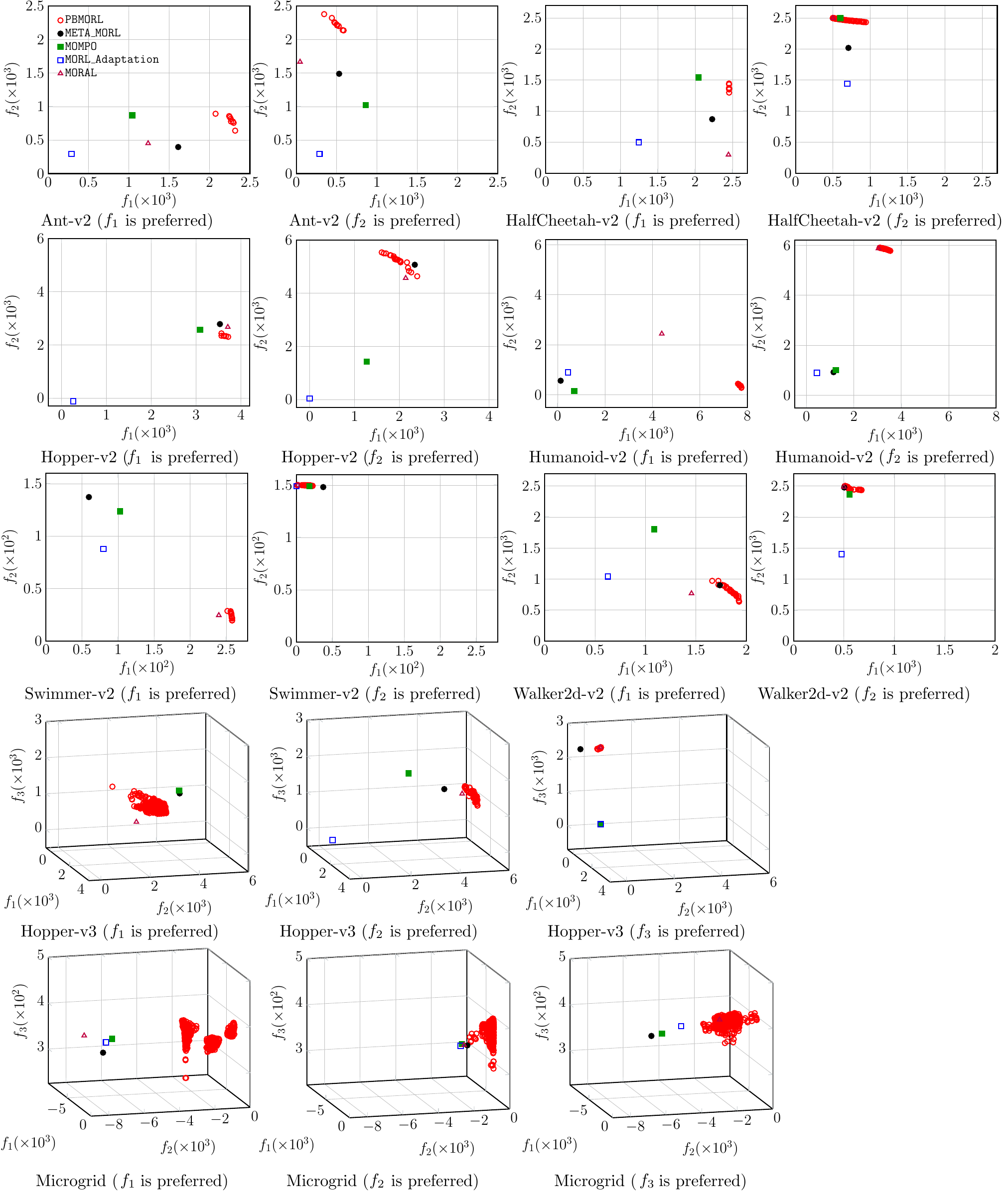}
    \caption{Plots of the non-dominated policies obtained by \our\ versus \texttt{MOMPO}, \texttt{META-MORL}, \texttt{MORL-Adaptation} and \texttt{MORAL} with different preferences on each objective.}
    \label{fig:appendix_exp2} 
\end{figure*}

\begin{figure*}[h]
	\center
    \includegraphics[width=\textwidth]{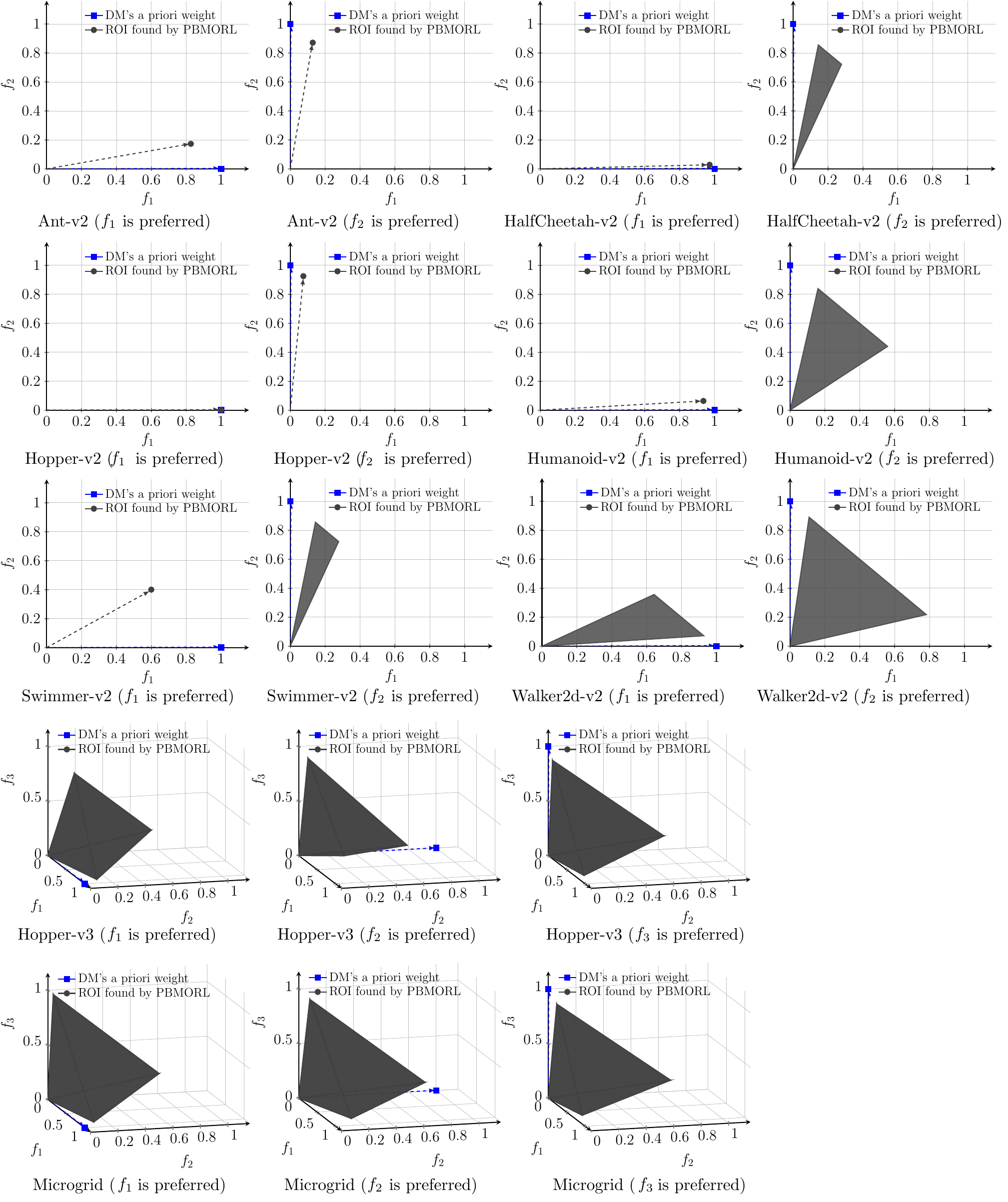}
    \caption{Comparison of the weight vector specified by the DM \textit{a priori} versus the ROI identified by \our\ (shaded in the gray region).}
    \label{fig:appendix_exp31} 
\end{figure*}

\begin{figure*}[h]
	\center
    \includegraphics[width=\textwidth]{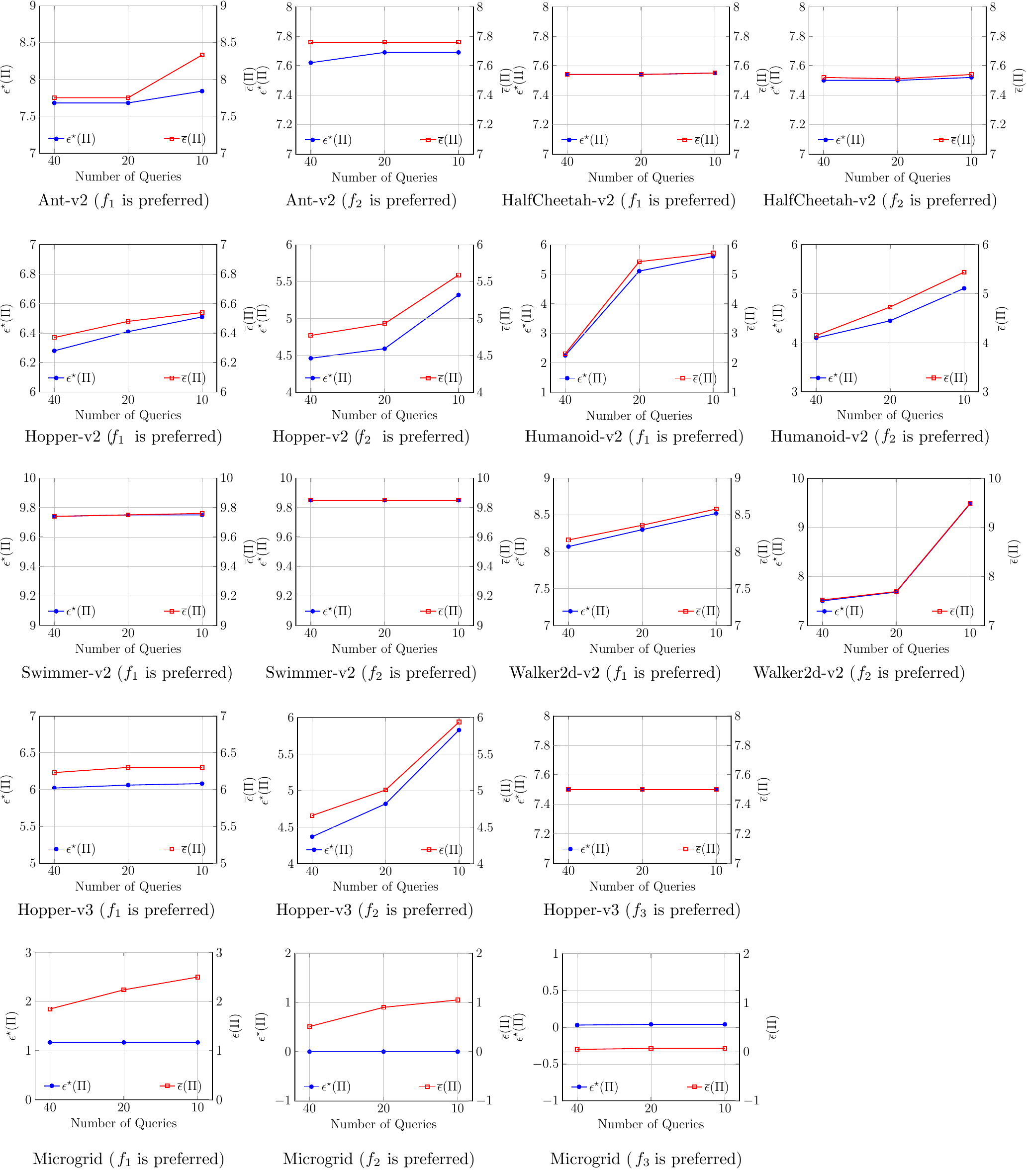}
    \caption{Comparison results of $\epsilon^\star(\Pi)$ and $\bar{\epsilon}(\Pi)$ obtained by \our\ with $10$, $20$, and $40$ interactions, respectively.}
    \label{fig:appendix_exp32} 
\end{figure*}